\documentclass[times,twocolumn,final,authoryear]{elsarticle}

\usepackage{framed,multirow}
\usepackage[top=0.6in, bottom=0.9in, left=0.65in, right=0.65in]{geometry} 

\usepackage{amssymb}
\usepackage{latexsym}

\usepackage{url}
\usepackage{xcolor}
\definecolor{newcolor}{rgb}{.8,.349,.1}
\usepackage{booktabs}
\usepackage{xtab,afterpage}
\usepackage{booktabs}
\usepackage{todonotes}
\usepackage{hyperref}

\newcommand{\quoten}[1]{``#1''}

\usepackage{makecell}

\journal{}

\makeatletter
\def\ps@pprintTitle{%
  \let\@oddhead\@empty
  \let\@evenhead\@empty
}
\makeatother

\begin{document}

\thispagestyle{empty}

% \ifpreprint
%   \setcounter{page}{1}
% \else
%   \setcounter{page}{1}
% \fi

% \begin{frontmatter}

%\clearpage
%\thispagestyle{empty}
%\ifpreprint
%  \vspace*{-1pc}
%\fi

%\begin{table*}[!th]
%\ifpreprint\else\vspace*%{-5pc}\fi

%\section*{Graphical Abstract (Optional)}
%To create your abstract, please type over the instructions in the template box below.  Fonts or abstract dimensions should not be changed or altered. 

%\vskip1pc
%\fbox{
%\begin{tabular}{p{.4\textwidth}p{.5\textwidth}}
%\bf Type the title of your article here  \\
%Author's names here \\[1pc]
%\includegraphics[width=.3\textwidth]{elsevier-logo}%%.pdf
%& 
%This is the dummy text for graphical abstract.
%}\\
%\end{tabular}
%}

%\end{table*}

\ifpreprint
  \setcounter{page}{1}
\else
  \setcounter{page}{1}
\fi

\begin{frontmatter}

\title{
\textbf{For a semiotic AI: Bridging computer vision and visual semiotics for computational observation of large scale facial image archives}
}

%\author[1]{Lia \snm{Morra}\corref{cor1}} 
%\ead{lia.morra@polito.it}
%\author[2]{Antonio \snm{Santangelo}\corref{cor2} \fnref{fn1}}
%\ead{antonio.santangelo@unito.it}

%\author[1]{Pietro \snm{Basci}}
%\author[1]{Luca \snm{Piano}}
%\author[1]{Fabio \snm{Garcea}}
%\author[1]{Fabrizio \snm{Lamberti}}
%\author[2]{Massimo \snm{Leone}}

\author[1]{Lia Morra \corref{cor1}} 
\ead{lia.morra@polito.it}
\author[2]{Antonio Santangelo \corref{cor2} \fnref{fn1}}
\ead{antonio.santangelo@unito.it}

\author[1]{Pietro Basci}
\author[1]{Luca Piano}
\author[1]{Fabio Garcea}
\author[1]{Fabrizio Lamberti}
\author[2]{Massimo Leone}

\fntext[fn1]{{All the authors collaborated in the theoretical discussion from which the draft of this article emerged. In particular, Antonio Santangelo wrote Sections 2.3 and 3, Lia Morra, Pietro Basci and Luca Piano the remaining sections.}}

\cortext[cor1]{Corresponding author: 
  Tel.: +0-000-000-0000;  
  fax: +0-000-000-0000;}
\address[1]{Politecnico di Torino, Corso Duca degli Abruzzi 24, 10129, Turin, Italy}
\address[2]{University of Turin, Via Verdi 8, 10124 Torino, Italy}

\newgeometry{width=\textwidth}
\leftskip=0.1\textwidth
\rightskip=0.1\textwidth
\begin{abstract} 
  Social networks are creating a digital world in which the cognitive, emotional, and pragmatic value of the imagery of human faces and bodies is arguably changing. However, researchers in the digital humanities are often ill-equipped to study these phenomena at scale. This work presents FRESCO (Face Representation in E-Societies through Computational Observation), a framework designed to explore the socio-cultural implications of images on social media platforms at scale. FRESCO deconstructs images into numerical and categorical variables using state-of-the-art computer vision techniques, aligning with the principles of visual semiotics. The framework analyzes images across three levels: the plastic level, encompassing fundamental visual features like lines and colors; the figurative level, representing specific entities or concepts; and the enunciation level, which focuses particularly on constructing the point of view of the spectator and observer. These levels are analyzed to discern deeper narrative layers within the imagery. Experimental validation confirms the reliability and utility of FRESCO, and we assess its consistency and precision across two public datasets. Subsequently, we introduce the FRESCO score, a metric derived from the framework's output that serves as a reliable measure of similarity in image content.
\end{abstract}
\restoregeometry

%\begin{keyword}
%  \MSC 41A05\sep 41A10\sep 65D05\sep 65D17
%  \KWD Keyword1\sep Keyword2\sep Keyword3

  %% MSC codes here, in the form: \MSC code \sep code
  %% or \MSC[2008] code \sep code (2000 is the default)
%\end{keyword}

\end{frontmatter}

%\linenumbers

%% main text
\section{Introduction}
\label{sec:intro}
In digital social networks, humans simultaneously produce and are exposed to an unprecedented amount of images. Many sociocultural practices are, as a consequence, changing the communicative power of digital representations and self-representations, most notably that of the human face. Digital image production has reached unprecedented levels in terms of quantity, pervasiveness, and potential for manipulation. The typical social media user spends more than two hours a day generating and scrolling through content, mostly in visual form \cite{owid-rise-of-social-media}. %https://datareportal.com/reports/digital-2023-october-global-statshot
Facebook, Instagram, Snapchat, Tinder, and other digital social networks are creating a digital world in which the cognitive, emotional, and pragmatic value of the imagery of human faces and bodies is arguably changing. However, researchers in the digital humanities are often ill-equipped to study these phenomena at scale. On the one hand, collecting and analyzing large amounts of images (so-called \emph{visual big data}) require semiautomatic tools and techniques for visualization, exploration, and tagging \citep{manovich2020cultural}. Although the analysis of textual media has progressed extensively, the analysis of visual media is lagging behind. Existing platforms do not cater to the needs of digital humanities or focus on low-level visual features \citep{ai4mediassh2022}. However, scholars in the digital humanities have developed sophisticated qualitative tools and techniques to interpret the multifaceted cultural significance of an image. There is a need to bridge these two approaches to reach insightful conclusions that are supported by adequate empirical evidence \citep{manovich2020cultural,ai4mediassh2022,berlanga2024digital}. 

Images on social media can be studied in many ways. In this article, we deal with the gaze we can cast on them, using the tools of visual semiotics \citep{polidoro2008che,eugeni2014analisi,pezzini2014corpi,mangano2018che,dondero2020linguaggi,corrain2023leggere}. We believe that this discipline asks itself a series of very general questions, the solution to which is the basis of the way in which all other disciplines, from psychology to sociology, from anthropology to aesthetics, from philosophy to art history, relate to this type of content. Visual semiotics, in fact, questions how we assign meaning to them, knowing full well that the interpretations we can produce are multiple. Nevertheless, it posits that any interpreter, when engaging with these forms of textuality, concentrates on certain specific fundamental components. These elements – of a plastic nature (shapes, colors, organization of space) or figurative (representations of the elements of the natural world), or related to the mechanisms which prompt the viewer to form a certain point of view on what is shown – are those that are usually considered pertinent by anyone who wants to assign a meaning to what they see in an image.

\begin{figure}
    \centering
    \includegraphics[width=\linewidth]{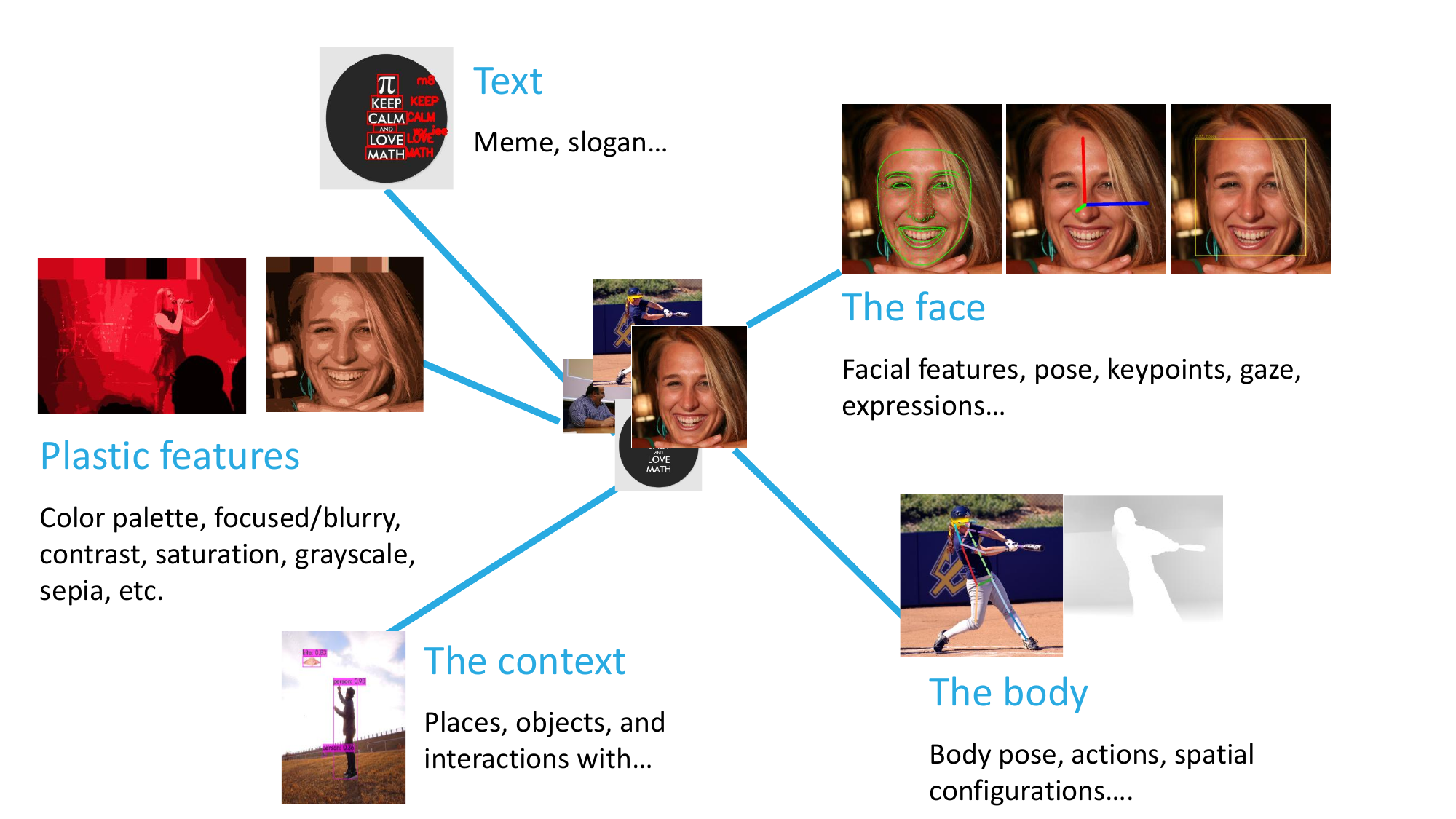}
    \caption{The FRESCO (Face Representation in E-Societies through Computational Observation) pipeline extracts quantifiable traits from images using SOTA computer vision and deep learning tools. The traits are not limited to facial and body characteristics, but encompass interaction with the context and background, the presence of textual elements, and so forth. Such traits are categorized according to their plastic (color, forms), figurative (objects and actions) and enunciative (gazes and mutual placements) categories or traits, based on principles from structural visual semiotics.}
    \label{fig:FRESCO overview}
\end{figure}

The core idea behind FRESCO (Face Representation in E-Societies through Computational Observation) was to develop a computational platform capable of bridging the gap between well-established semiotics principles and quantitative computational image interpretation techniques that could scale to hundreds and thousands of images. It builds on the tremendous advances in computer vision (CV) over the past decades and recognizes the potential of both established image processing techniques and the most recent foundational models in extracting \textit{traits} from images, that is, characteristics that would be considered as potentially \textit{pertinent} by visual semiotics scholars. As an example, and without loss of generality, such a platform could be used to cluster images produced by social media users based not only on their content, but also on their composition or their narrative structure. The FRESCO platform, by deconstructing images in a series of numerical and categorical variables, as depicted in Figure \ref{fig:FRESCO overview}, enables semioticians to take advantage of the extensive toolbox that the field of big data analytics and data mining has developed in the last decades to uncover novel and unexpected patterns from large visual collections.

In synthesis, our contributions are as follows:

\begin{itemize}
    \item we introduce FRESCO, a computational framework that operationalizes  structural visual semiotics  in order to investigate the socio-cultural meaning of social media images at scale;
    \item we propose a practical implementation of the FRESCO framework and experimentally validate it on human-centered datasets to demonstrate the validity and usefulness of the proposed framework;
    \item we propose the FRESCO-score, a principled and transparent similarity measure based on the output of the FRESCO pipeline. 
\end{itemize}

The remainder of the paper is organized as follows. Section \ref{sec:related} presents an overview of the related work. Section \ref{sec:background} provides some essential background on visual semiotics, while Section \ref{sec:fresco} illustrates the FRESCO computational pipeline in detail.  Sections \ref{sec:experimental} and \ref{sec:results} present the experimental methodology and results, which are discussed in Section \ref{sec:discussion}. Finally, Section \ref{sec:conclusions} concludes the paper and suggests future work.

\section{Related work}
\label{sec:related}
Many authors have investigated the interplay between computer vision and disciplines from the humanities, in particular between computer vision and art/media analysis \citep{datta2006studying,hussain2017automatic,ye2018advise,madhu2020understanding,wijntjes2021shadows,stork2021computational,santos2021artificial,arnold2022automatic,yi2023towards}, psychology \citep{strano2008user,ferwerda2015predicting,vilnai2015picture,segalin2017social,segalin2017your,cucurull2018deep,branz2020red} and semiotics \citep{reyes2019new,ghidoli2021finding,pandiani2023seeing}. In this section, the most relevant works to FRESCO and social media analysis in general are briefly reviewed. 

\subsection{Inferring personality from social media}
Some studies show that it is possible, to some extent, to infer psychological traits from images published on social media, such as profile pictures \citep{branz2020red,ferwerda2015predicting,cucurull2018deep,segalin2017social,segalin2017your,vilnai2015picture,strano2008user}. For instance, \citet{segalin2017social,segalin2017your} investigated the ability of hand-crafted features and deep learning to infer self-assessed and attributed personality traits based on image features extracted from Facebook profile pictures. Their research suggests that images associated with a person can reveal some of their individual characteristics, such as their personality traits, with computerized assessment even outperforming human evaluation \citep{segalin2017social}.  In this type of study, social media users can be subjected to online questionnaires designed to self-assess personality traits, and then the classifiers are trained to predict the labels extracted from the questionnaire. The task is well defined with clear labels, and the problem is to extract/select relevant information. In FRESCO, we do not wish to make predictions on individual social media users, but we are rather interested in extracting multi-faceted, culturally relevant aspects of digital imagery. 

% art history and photography 
\subsection{Computational analysis in media and art history}
\label{ssec:media_art_history}

Several computational platforms have been developed to analyze image archives in art history
\citep{madhu2020understanding,yi2023towards,wijntjes2021shadows,stork2021computational,chen2015artistic,seguin2016visual,elgammal2018shape}, artistic/historical photography \citep{datta2006studying,arnold2022automatic,mannisto2022automatic,arnold2020enriching} and advertisement \citep{ye2018advise}. Computerized tools can analyze, at scale and in a systematic fashion, large image archives. For instance, \citet{elgammal2018shape} showed how machine learning can predict styles based on visual features and relate them to art history concepts. They show that representations learned by deep learning correlate with principles from art history and that predictions align with historical progression, thus providing a quantifiable verification of art historical theories. Other authors have focused on the use of machine learning to model and quantify image composition \citep{chen2015artistic} or image aesthetics \citep{yi2023towards}. Computerized analysis also allows art historians to establish links between different authors or artworks that may otherwise go undetected \citep{seguin2016visual}.

While most of the above mentioned studies have focused on a few variables or on a specific analysis, in more recent years scholars have started to suggest that, in light of recent advances in computer vision and deep learning, a more extensive \quoten{visual grammar} could be operationalized and made accessible to the digital humanities scholar. In particular, \citet{mannisto2022automatic} have proposed the AICE framework (Automatic Image Content Extraction) tailored to photography analysis. Their framework is based on the theoretical underpinnings of visual semiotics, and in particular the book \quoten{Image Grammar of Visual Design} by \citet{kress1996reading}, and \citet{bell2012content}'s version Visual Content Analysis (VCA), a more practical and readily operable adaptation of the original grammar which is particularly suitable for photographic analysis.  
In their book, \citet{kress1996reading} presented an inventory of the major composition structures established as conventions in the history of visual semiotics and examined how they are used by contemporary image makers to generate meaning. Despite being developed independently and from different sources, FRESCO and AICE share many common characteristics. Both methods share the premises that visual semiotics provides a theoretical background to define a comprehensive lists of variables, which are mapped to state of the art computer vision and machine learning techniques. FRESCO is structured differently, grouping concepts according to different levels of analysis (plastic, figurative, and enunciational) initially defined by Greimas and refined by subsequent authors, as presented in greater detail in Section \ref{sec:background}. We carefully reviewed the structure proposed in AICE to ensure that all the variables proposed therein are also covered in FRESCO. In addition, unlike \citet{mannisto2022automatic} we provide a first practical implementation of FRESCO and go into greater detail into the accuracy and consistency of the extracted values, as well as practical issues that arise when trying to combine them into an overall similarity score.

\subsection{Semiotics and computational analysis}

Regarding the relationship between semiotics and computational analysis, the debate has first and foremost focused on how a discipline that originated at the intersection of philosophy and the social sciences, thus in the humanities, can dialogue with computer science and statistics. In this regard, a very important book is \textit{Quantitative Semiotic Analysis} \citep{compagno2018quantitative}, in which the issue of how to use tools for quantitative investigation is addressed when we set out to identify the meaning of a written or visual text, an activity that in the past has always been carried out using qualitative analysis methodologies. 

Since signification is a deferential phenomenon, which people accomplish by linking signs to what those signs mean, thanks to codes that are not found in the texts themselves but in the minds of the interpreters and the culture they share, it has always proved more functional to assign the task of describing these mechanisms to a researcher and his or her ability to produce interpretations, as well as to imagine or recognize the interpretive logics of others. However, this inevitably reduces the extent of the content corpora that can be worked on, since this kind of investigation is constrained by methods of analysis that take a long time to be carried out. When one wishes to conduct studies on a very large corpus of texts, such as can be built in digital environments, it is necessary to make use of quantitative methods and tools. Confronting the various approaches to this problem in the various fields of the digital humanities \citep{moretti2005graphs,manovich2020cultural}, the authors of \textit{Quantitative Semiotic Analysis} propose solutions that are in many ways similar to those we adopt in FRESCO: they emphasize, in fact, that computer systems must be designed to make use of semiotics to scrutinize their objects of analysis, recognizing their most significant elements and describing them in a way that is as functional as possible to enable the researchers who use them to best interpret the value of the data that these same systems produce.    

The book edited by \citep{compagno2018quantitative}, however, is also interesting for another reason: it deals, in fact, with a long series of theoretical problems raised by the encounter between semiotics and the techniques of quantitative investigation of large digital data corpora, but in its most applied part it deals only with the analysis of written texts. Only one article - that of Cholet (\citep{cholet2018images}, ibid.: 101-121) - deals with the study of images, but with the technique of eye tracking. Thus, in practice, FRESCO's field of research is not considered, in this work. As is well known, after all, the computer tools that are used today to carry out quantitative semiotic analysis in the digital domain are mostly linguistic systems. So far, little has been done to reason about how to read and process images in an automated way using semiotics.

In this regard, the scarce available literature can be found first of all in the field of marketing. \citet{ghidoli2021finding} reflect on some computer tools used to identify trends in consumer tastes. Using the Java SOM Toolbox framework, for example, \citet{o2015multimodal} analyzed large masses of images found online of young Japanese people having their pictures taken in their favorite clothes, producing a graph that can show how these can be divided into interrelated classes according to some logic that takes into account how fashion works in those latitudes \citep{owyong2009clothing}. Something similar, but at a broader level of generality, was done by the authors of ScenarioDNA\footnote{https://www.scenariodna.com/}. In this case, different types of images found on social networks have been organized into concept maps, which allow them to be grouped into clusters of similar content, which derive their meaning because they differ from those found in other clusters that can be linked to them. By doing so, it is possible to conduct synchronic and diachronic analyses of the spread of these same contents. In addition, thanks to some network analysis tools, it is possible to understand how certain images spread in some networks of people rather than others. None of these systems focus, as FRESCO does, on face analysis, but the fact that they are beginning to be developed demonstrates the significance of our research project.

Another research field in which semiotics and computer science have often intersected is that of media and art history, as detailed in previous Section \ref{ssec:media_art_history}. Among those, the AICE framework for the analysis of photographic archives\citep{mannisto2022automatic}, or the Distant Viewing tookit proposed by \citep{arnold2019distant},  are heavily based on semiotic principles. Our proposal pushes past these previous attempts by providing and validating an incomplete but extensive implementation of the proposed concepts.   

Other works have relied on ideas that can be traced back to semiotics to design novel tasks for the computer vision community, focusing on the interpretation of ``higher-level'' semantic information or abstract concepts from still images \citep{pandiani2023seeing,martinez2024wicked}. For instance, \citet{tores2024visual} proposed a computer vision task to detect the ``male gaze'' from video, that is, the objectification of women in video based on multiple cues such as camera placement and movement, gaze interactions, choice of clothing or nudity, posture, etc., many of which are also present in FRESCO. Other works have focused instead on quantifying image compositions in artworks based on pose and gaze information, focusing not only on subjects' pose, but also on composition lines established by the subjects' gazes \citep{madhu2020understanding}. 

\color{black}

\section{Background}
\label{sec:background}
As we have anticipated, FRESCO has been designed to allow scholars of visual semiotics to analyze the meaning of large amounts of images taken from the social profiles of people all over the world. Since the interpretation of this kind of content can differ depending on the research questions and the point of view of the researcher, our goal was to develop a computer system capable of reading the constituent elements of the images themselves which, according to the scientific literature, are usually taken into account to determine the meaning of the latter,  whatever it is.

To identify these elements, we have used several texts, starting with Greimas' seminal essay entitled \textit{Sémiotique figurative et sémiotique plastique} \citep{greimas1984semiotique}, cited by many as the foundational work of modern visual semiotics studies \citep{corrain2023leggere}. Then we turned to books on semiotic analysis of visual text in general \citep{polidoro2008che,eugeni2014analisi,dondero2020linguaggi}. Finally, we consulted works that deal with the semiotic study of photography \citep{mangano2018che} and images on social media \citep{pezzini2014corpi}. 

All the authors of these articles and volumes agree that when we are faced with a figurative image such as those that, in most cases, are uploaded to our social profiles by people, one of the first interpretative actions is the recognition of the figures of the natural world that it reproduces: humans, animals, plants, objects, places, etc. It is also essential to recognize the actions of these subjects, which of them are active and which are passive, how they move, and what emotions they feel. All this serves to identify the main topic or topics of this image, but to do so and understand how the image frames the topic itself, it is also necessary to focus on the so-called \quoten{plastic} level. The latter comprises three categories of traits: eidetic, chromatic, and topological. The first category (eidetic) accounts for the shapes, lines, contours, dimensions, and symmetries of which the image is composed. The second (chromatic) for the colors, brightness, saturation, and textures. The last one (topological) for the spatial arrangement of all these contents, that is, what is above or below, right or left, in the center or in the periphery, in the foreground or in the background. All these elements, which compose the plastic structure of the image, contribute, together with the more figurative ones, to determine its meaning.

\begin{figure}
     \centering
     \includegraphics{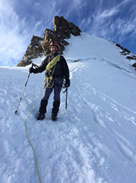}
     \caption{The profile of a mountain climber}
     \label{fig:example1}
 \end{figure}

For example, as we have shown in a previous work \citep{santangelo2023}, in order to understand the meaning of Figure \ref{fig:example1}, downloaded from the Facebook/Meta profile of one of the authors of this article, it is certainly important to understand that it depicts a man with mountaineering equipment, a mountain peak to climb, and a very steep slope made of snow and ice. But it is also essential to realize that the mountaineer covers only a small part of the image itself, which is otherwise occupied by the majesty of the natural environment; that he is more or less in the middle of the frame; that at the top of the image is the mountain top from which he has descended or on which he will soon climb, while a steep slope lies below; that the light illuminates his smiling face, giving it a warm hue in a context otherwise populated by cold colors. All these figurative and plastic elements help to communicate the happiness of being in the beauty of wild nature and being able to climb, feeling small but at the same time being the protagonist of a great adventure.
 
\begin{figure}[!tb]
     \centering
     \includegraphics{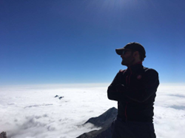}
     \caption{Picture of a man looking towards the clouds beneath him}
     \label{fig:examples2}
 \end{figure}

Another fundamental element, in order to understand the meaning of Figure \ref{fig:example1}, as of any other image, is the construction of the observer's point of view, which \citet{eugeni2014analisi} (op. cit.: 97-166) also calls \textit{gaze system} or \textit{watcher-looked system}. Eugeni himself argues that, depending on whether the latter is \textit{basic}, \textit{first-grade} or \textit{second-grade}, it helps position the viewer of the image with respect to the latter and its contents, guiding how the image is \quoten{read} by the viewer. For example, speaking of the basic watcher-looked system, it is evident that a large painting of the face of Christ on the dome of a church is meant to be observed by much smaller people who are below, giving it a very specific meaning. However, a photograph on a social network page is composed to be observed from a very different position, which also generates greater engagement due to its communication style. On the other hand, speaking of the first-degree watcher-looked system, which again in photography refers substantially to the way in which the camera (or the camera of a smartphone) is placed, if we pay attention again to Figure \ref{fig:example1}, the fact that it is taken from below and from a distance puts the observer in a position to appreciate the great steepness of the slope and the vertigo of the climb. Finally, coming to the \textit{second-degree watcher-looked system}, which has to do with the direction of the gazes of the subjects represented, Figure \ref{fig:examples2}, also downloaded from the Facebook/Meta profile of one of the authors of this article, shows how important it is to look in the direction in which the protagonist of an image is looking, since there, evidently, lies a good part of the meaning of what the image itself wants to communicate.

\begin{figure}
     \centering
     \includegraphics[width=0.15\linewidth]{images/Picture1.png}
     \includegraphics[width=0.25\linewidth]{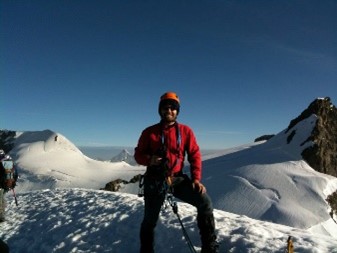}
     \includegraphics[width=0.25\linewidth]{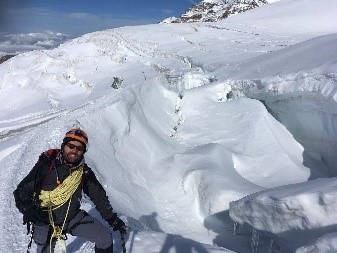}
     \includegraphics[width=0.25\linewidth]{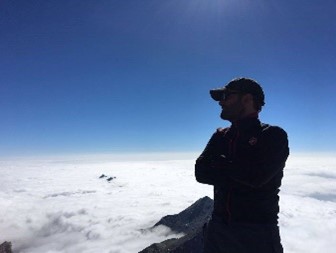}
     \caption{A set of images with similar meaning}
     \label{fig:example3}
 \end{figure}

A system like FRESCO must be able to recognize all the salient characteristics of a plastic, figurative nature and related to the construction of the gaze of the observer of an image, in order to combine them with those of the other images it processes. Such a system should aid the researcher expert in visual semiotics in identifying clusters of images, such as the three ones depicted in Figure \ref{fig:example3}, that have many similar elements within them and, therefore, can be interpreted in the same way.

\section{The FRESCO architecture}
\label{sec:fresco}
\subsection{Conceptual design}
\label{sec:fresco-design}

The FRESCO architecture arises from a systematic mapping activity between concepts introduced in visual semiotics, introduced for the uninitiated reader in Section \ref{sec:background}, and concepts and techniques developed in the context of CV. This mapping is informed by the authors' experience, by an extensive analysis of the current literature, as well as previous attempts from art history and photography history, such as AICE \citep{mannisto2022automatic}. The result is presented in Table \ref{tab:fresco}, in which the first two columns refer to
the traits or categories commonly used in visual semiotics, while the last three columns denote their CV counterparts. Each trait was associated with one or more CV tasks that compute one or more quantitative measure: when this column is empty, it does not necessarily imply that the corresponding trait is not 
amenable to computerized analysis, but rather that to the best of the authors' knowledge the task has not been extensively tackled in the literature, and therefore suitable annotated datasets and models are not available. The last column illustrates the numeric output that is used to quantify the corresponding trait. In some cases, the output of a CV algorithm or model could be another image, such as a semantic segmentation map. To enable certain types of analysis, it would be preferable to have more synthetic and numeric measures: for instance, if an image is reduced to a series of numeric or categorical measurements, it makes it easier to apply data analytics techniques to correlate them with other variables representing, e.g., socio-demographic measurements. For the traits currently implemented in FRESCO.v1, we therefore sought to define measurements that could be used for this purpose.

%\begin{center}
%\begin{longtable}{cp{3cm}|p{2cm}p{2cm}p{2cm}}

\afterpage{\onecolumn

%\begin{landscape}

\topcaption{Face Representation in E-Societies through Computational Observation (FRESCO) Computational Framework. The table maps semiotic-inspired category or variable with the corresponding computer vision task, if available, and the values that the variable can assume (either output by the computer vision task, or appropriately summarized). A \checkmark indicates variables that are included in the current FRESCOv1 implementation and the experimental validation in this paper. } \label{tab:fresco}
\tablefirsthead{    \hline	&	\textbf{Semiotic category}	&	\textbf{Computer vision task}	&	&	\textbf{Values}	\\ \hline}
\tablehead{\multicolumn{5}{c}{Table \ref{tab:fresco}, cont'd}\\}
\tablelasttail{\bottomrule}

\begin{xtabular*}{\linewidth}{p{0.8cm}p{4cm}|p{4cm}p{0.5cm}p{5cm}}
	&		&		&		&		\\
	&		&		&		&		\\   		
0	&	\textbf{{Technical}} &	&				&		\\
0.1	&	Image type	&	classification / clustering	&	\checkmark	&	Photograph / illustration / map / ..	\\
	&		&		&		&		\\

\hline \\
1	&	\textbf{{Plastic level	}} &		&		&		\\

1.1	&	\textbf{Eidetic categories}	&		&		&		\\
1.1.1	&	Form	&		&		&		\\
1.1.2	&	Simmetry	&		&		&		\\
1.1.3	&	Mimetic/abstract	&		&		&		\\
1.1.4	&	Geometric/non-geometric	&		&		&		\\
1.1.5	&	Kind of contour 	&		&		&		\\
1.1.6	&	Lines	&	Edge extraction	&	\checkmark	&	Line map	\\
& & Diagonal element detection  &  &  diagonal\_ulbr, diagonal\_urlb, ... \\

	&		&		&		&		\\
1.2	&	\textbf{Chromatic categories}	&		&		&		\\
1.2.1	&	Color	&	Palette estimation 	&	\checkmark	&	Palette	\\
	&		&	Color 	&	\checkmark	&	Grayscale / color 	\\
	&		&	Color distribution 	&	\checkmark	&	Histogram	\\
1.2.2	&	Luminosity 	&	Brightness estimation	&	\checkmark	&	Brightness	\\
1.2.3	&	Saturation	&	Saturation estimation	&	\checkmark	&	Saturation 	\\
1.2.4	&	Texture 	&	Texture classification 	&		&		\\
	&		&		&		&		\\
	&		&		&		&		\\
1.3	&	\textbf{Topological categories}	&		&		&		\\
1.3.1	&	High/low	&	Object detection	&	\checkmark	&	Position of each object centroid w.r.t. the image midline	\\
1.3.2	&	Left/right 	&	object detection	&	\checkmark	&	Position of each centroid w.r.t. the image midline	\\
1.3.3	&	Central/peripheral	&	object detection	&	\checkmark	&	Centrality ratio of each object	\\
1.3.4	&	Foreground/background	&	panoptic segm.  + depth	&	\checkmark	&	Avg. depth value of each person	\\
	&		&	panoptic segm.  + depth&	\checkmark	&	Avg. depth value of each object	\\
1.3.5	&	Spatial disposition of forms	&	visual relationship detection	&		&	scene graph ('X' left of 'Y', etc.)	\\
	&		&	semantic segmentation	&	\checkmark	&	semantic segmentation map	\\
	&		&	semantic segmentation	&	\checkmark	&	spatial coverage (percentage covered by each class)	\\
  & & spatial composition class & & vertical, horizontal, centered \\
1.3.7	&	Dynamization of forms	&		&		&		\\
	&		&		&		&		\\
1.4	&	Links between adjacent forms	&		&		&		\\
1.4.1	&	By similarity	&		&		&		\\
1.4.2	&	By confrontation	&		&		&		\\
1.5	&	Links between distant forms	&		&		&		\\
1.5.1	&	By similarity	&		&		&		\\
1.5.2	&	By confrontation	&		&		&		\\
	&		&		&		&		\\
1.6	&	Overall configuration	&		&		&		\\
1.6.1	&	Static vs. Dynamic	&	Classification 	&		&		\\
	&		&		&		&		\\
\hline \\
2	&	\textbf{Figurative level}	&		&		&		\\
2.1	&	\textbf{General}	&		&		&		\\
2.1.1	&	Main topic	&	Classification / Clustering	&	\checkmark	&	Person/animal/object/environment/event/...	\\
	&		&	Image tagging	&	\checkmark	&	tags 	\\
2.1.2	&	Salience	&	Salience estimation	&		&	salience map	\\
	&		&		&		&		\\
2.2	&	\textbf{Persons / objects / scene }	&		&		&		\\
	&		&		&		&		\\
2.2.1	&	\textbf{Characteristics of people groups} 	&		&		&		\\
2.2.1.1	&	Number of people	&	Face/person detection	&	\checkmark	&	0/1(single)/2(couple)/3-6(small group)/7-12(medium group)/13-30(large group)/31-(crowd)	\\
2.2.1.2	&	Number of groups	&	Gaze estimation / Social distance estimation	&		&	1/2/3+	\\
2.2.1.3	&	Group typology	&	Gaze estimation / Social distance estimation	&		&	Unfocused/ common focused / jointly focused/	\\
2.2.1.4	&	Group type	&	Classification	&		&	Family/friends/sport team/...	\\
2.2.1.5	&	Atmosphere	&	Classification	&		&	Casual/formal/intimate/festive/...	\\
	&		&		&		&		\\
	&		&		&		&		\\
2.2.2	&	\textbf{Characteristics of each person	}&		&		&		\\
2.2.2.1	&	Status	&	Main character recognition	&		&	Main Character (MC) / Side Character (SC)	\\
2.2.2.2	&	Age	&	Age estimation	&	\checkmark	&	Baby/child/young/adult/old	\\
2.2.2.3	&	Gender	&	Attribute prediction	&	\checkmark	&	Male/female/other	\\
2.2.2.4	&	Identity	&	Face recognition	&		&	Name	\\
2.2.2.5	&	Ethnicity	&	Attribute prediction	&	\checkmark	&	European/Asian/African/...	\\
2.2.2.6	&	Height	&	Height estimation	&		&	Short/average/tall	\\
2.2.2.7	&	Weight	&	Weight estimation	&		&	Thin/average/fat	\\
2.2.2.8	&	Occupation	&	Classification	&		&	Doctor/police/cook/pilot/	\\
2.2.2.9	&	Role	&	Visual relationship detection	&		&	Child/mother/friend/neighbor/...	\\
2.2.2.10	&	Nudity	&	Nudity detection	&		&	Nude/partially nude/clothed	\\
2.2.2.11	&	Physical condition	&	classification	&		&	Healthy/sick/wounded/dead/...	\\
2.2.2.12	&	Clothes	&	Attribute classification / Object detection	&	\checkmark	&		\\
2.2.2.13	&	Clothing style	&	Style clustering	&		&		\\
2.2.2.14	&	Face and head accessories	&	Attribute prediction	&	\checkmark	&	Glasses/jewellery/hat/..	\\
2.2.2.15	&	Facial attributes	&	Attribute prediction	&	\checkmark	&	Eyebrows/nose type/double chin/cheeks ...	\\
2.2.2.16	&	Facial expressions	&	Attribute prediction	&	\checkmark	&	Smile/frown/...	\\
	&		&	Facial keypoints	&		&		\\
2.2.2.17	&	Hair attributes	&		&	\checkmark	&	Beard /Hair length/Hairline/Bangs/Sideburns...	\\
	&		&		&		&		\\
2.2.3	&	\textbf{Objects}	&		&		&		\\
2.2.3.1	&	Status	&		&		&	Main motif (MM)/side motif (SM)	\\
2.2.3.2	&	Category	&	Object recognition	&	\checkmark	&	Animal/object	\\
2.2.3.3	&	Text in image	&	Text recognition / OCR	&	\checkmark	&	Text in image	\\
	&		&		&		&		\\
2.2.4	&	\textbf{Settings/events}	&		&		&		\\
2.2.4.1	&	Scene class	&	Scene classification / Tagging	&	\checkmark	&	 Urban/rural/forest/hospital/school/	\\
2.2.4.2	&	Privacy	&		&		&	Private/semi-public/public	\\
2.2.4.3	&	Indoor/outdoor	&	scene classification	&	\checkmark	&	Indoor/outdoor	\\
2.2.4.4	&	Man-made/natural	&	scene classification	&	\checkmark	&	Man-made/natural	\\
2.2.4.5	&	Event	&	event recognition 	&		&		\\
2.2.4.6	&	Location	&	location recognition / landmark detection	&		&	Location	\\
2.2.4.7	&	Time of day	&	classification	&		&	Morning/day/evening/night	\\
2.2.4.8	&	Time of year	&	classification	&		&	Winter/spring/summer/autumn	\\
2.2.4.9	&	Weather	&	classification	&		&	Sunny/cloudy/raining/snowing/...	\\
	&		&		&		&		\\
2.3	&	\textbf{Movement}	&		&		&		\\
2.3.1	 &	Type of movement &		&		&	Blocked / contracted / articulated		\\
2.3.2	&	Visibility	&		&		&		Hidden / manifest\\
	&		&		&		&		\\
2.4	&	\textbf{Action}	&		&		&		\\
2.4.1	&	Single action	&	action recognition 	&		&	classification	\\
	&		&	body pose	&	\checkmark	&	pose	\\
	&		&	caption generation 	&	\checkmark	&	textual description	\\
2.4.2	&	Aggregate of actions	&	 visual relationship detection	&		&	scene graph	\\
2.4.3	&	Narrative	&		&		&		\\
	&		&		&		&		\\
2.5	&	\textbf{Emotions}	&		&		&		\\
2.5.1	&	Intensity	&	Arousal regression	&	\checkmark	&	arousal	\\
2.5.2	&	Emotion recognition	&	emotion classification	&	\checkmark	&	happy / neutral / fear / sadness / disgust	\\
2.5.3	&	Emotional valence	&	valence regression	& \checkmark	& valence			\\
	&		&		&		&		\\

\hline \\
3	&	\textbf{{Enunciational level}}	&		&		&		\\
3.1	&	\textbf{Basic watcher-looked system}: the viewer	&		&		&		\\
3.1.1	&	position of the viewer	&	panoptic segm.  + depth	&	\checkmark	&	distance of the main subject(s) from the camera	\\
3.1.2	&	position of the viewer	&	panoptic segm.  + depth	&	\checkmark	&	distance of the main character(s) from the camera	\\
3.1.3	&	position of the viewer	&	horizon line estimation	&		&	position of the horizon line  (frontal/from above/from below)	\\
&  &	scene classification 	&	\checkmark	&	indoor/outdoor	\\
& &		framing	&	\checkmark	&	portrait vs. scene	\\
3.1.4	&	position of the camera	&	camera pose estimation 	&		&		\\
	&		&		&		&		\\
3.2	&	\textbf{First-grade secondary watcher-looked system}: the observer subject 	&		&		&		\\
3.2.1	&	position of the observer	&	head pose	&	\checkmark	&	angle (yaw/pitch/roll)	\\
3.2.2	&	position of the observer	&	body pose	&	\checkmark	&	shoulder/hip angle (frontal/rotated left/rotated right)	\\
3.2.3	&	position of observer	&	gaze direction	&	\checkmark	&	angle (yaw/pitch)	\\
3.2.4	&	position of the observer	&	presence/absence of perspective	&	&	classification	\\
3.2.5	&	position of the observer	&	vanishing point regression	&		&	vanishing point positions wrt the image frame	\\
	&		&		&		&		\\
3.3	&	\textbf{Second-grade secondary watcher-looked system}: indicators/bystanders; insignias and epigraphs	&		&		&		\\
%3.3.1	&	bystanders  (subjects who watch the main scene)	&	main character detection + gaze  detection	&		&		\\
%3.3.2	&	indicators (subject who watch the most important action in the scene)	&	main character detection + action recognition + pose estimation	&		&		\\
3.3.1	&	bystanders 	&	main character detection + gaze  detection	&		&		\\
3.3.2	&	indicators 	&	main character detection + action recognition + pose estimation	&		&		\\

3.3.3	&	insignias 	&	object detection	&		&		\\
3.3.4	&	epigraphs	&		&		& \\	
3.4	&	\textbf{Spatial relationship}	&		&		&		\\
3.4.1	&	of secondary watcher-looked systems (first and second grade)	&		&		&		\\
3.4.1	&	of secondary watcher-looked systems (first and second grade) vs. basic watcher-looked system	&		&		&	coincident / rotated left / rotated right / opposite	\\
3.4.2	&	of first grade secondary watcher-looked system vs. second grade secondary watcher-looked system	&		&		&		\\

    \hline
\end{xtabular*}

%\end{landscape}
\twocolumn
} % end of scope of "\afterpage" directive

The first section of Table \ref{tab:fresco} represents technical information regarding the nature of the image: a photograph, an illustration, a map, a drawing, etc. The nature of this classification depends in part on the assumptions made about the archive under analysis. FRESCO was initially designed for the analysis of social media profile pictures, and thus to accommodate any type of imagery that a user may potentially select as a symbolic depiction of their face, while maintaining a focus on the face. Classifiers to distinguish different types of mediums can be trained with high accuracy \citep{cutzu2003estimating,wevers2020visual}. 
For example, \citet{wevers2020visual} trained a CNN to distinguish historical photographs from several types of diagram. Alternatively, and without the need to define ad hoc categories, one can obtain a rough classification by employing a clustering technique on the features extracted from a pre-trained model.  In the case of photographs, technical characteristics are often available from the file header, such as camera model, focal length, etc. (see \citet{mannisto2022automatic} for a more thorough analysis of this aspect). 

% primo paragrafo: caratteristiche plastiche
At the \textit{plastic level},  the meaning of an image is constructed through a complex interplay of eidetic, chromatic, and topological categories. These plastic elements not only accentuate, but also sometimes contradict the figurative content, leading to nuanced interpretations and visual ambiguities. Many of the plastic categories identified in visual semiotics correspond to low-level image characteristics that have been studied in image processing and computer vision for decades.

\textit{Eidetic} categories (1.1) pertain to the forms expressed in the images, through lines, contours, and textures. As discussed in \citet{eugeni2014analisi}, eidetic categories include first of all whether the image has mimetic or abstract qualities - that is, whether the image seeks to represent an existing objects or is rather an abstract image. Eidetic categories properties of the overall spatial composition such as the type of forms present (circular, square, etc.) (1.1.1), the symmetry of the composition and the main objects (1.1.2), the type of contours present (1.1.5), the main lines forming in the composition (1.1.6), etc. Many CV techniques have been developed to characterize the overall spatial composition of an image 
\citep{yao2012oscar,amirshahi2014evaluating,wevers2020visual}, such as evaluating the rule of thirds \citep{amirshahi2014evaluating}. The presence of prominent compositional elements, such as diagonal line detection (1.1.7) and classification of spatial composition as vertical, horizontal or central (1.3.5), can be established borrowing from the field of computational photography \citep{yao2012oscar}.

In particular, spatial composition is of particular importance in the study of artistic photography \citep{yao2012oscar}, advertising \citep{wevers2020visual}, and paintings \citep{dondero2020linguaggi}, in which the author of the image usually employs more sophisticated control over the composition of the image. In FRESCO.v1, we include only edge extraction among the existing tools. 

\textit{Chromatic} features encompass color (1.2.1), luminosity (1.2.2), saturation (1.2.3), and contrast, influencing emotional resonance and symbolic associations within the image. Chromatic features in FRESCO.v1 include global image features, such as palette and color histogram. Textural components, such as texture classification and clustering of image pixels based on textural and chromatic components \citep{bianconi2021colour} will be included in future work.

\textit{Topological} features refer to spatial relationships, perspective and arrangement of elements, shaping the overall composition. Spatial relationships can be inferred from CV tasks such as object detection, semantic segmentation, panoptic segmentation, depth estimation and visual relationship detection. These tools produce as output spatial maps that can be directly used to, e.g., search for images with similar composition in terms of segmentation or depth map. However, as stated before, we sought to define more concise and interpretable quantities to enable efficient indexing and comparison of large-scale image collections. 

\begin{figure}[!tbh]
   \centering
   \includegraphics[width=\linewidth]{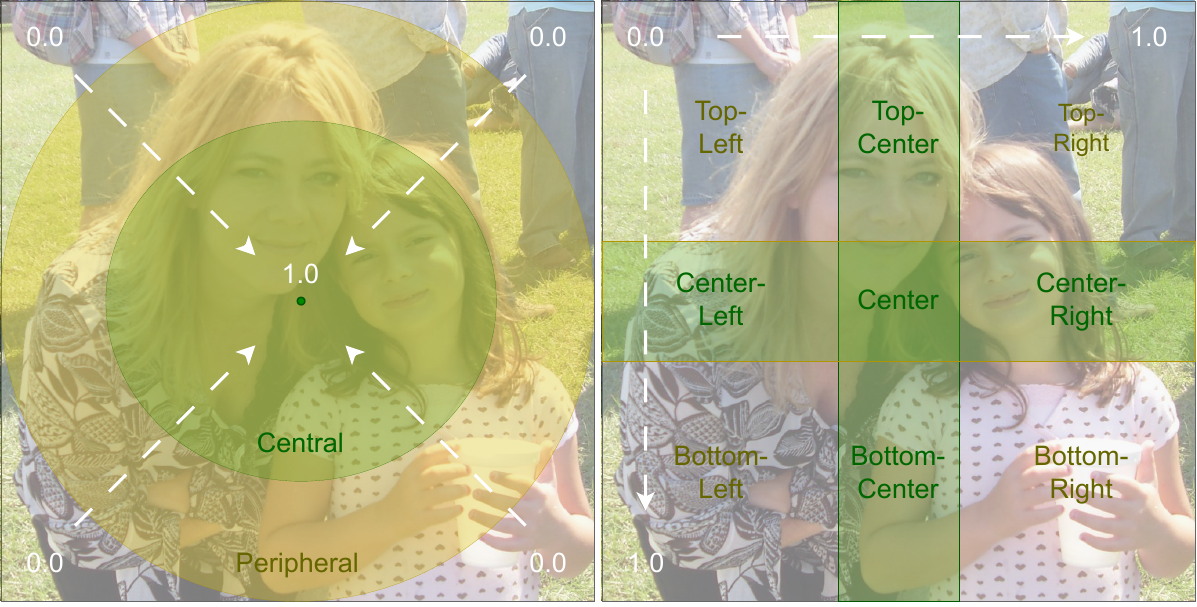}
   \caption{We compute position of the centroid of each identified object or person with respect to the vertical and horizontal midlines, as well as the distance from the image center, to determine the position of each object or person with respect to the image frame. All positions are rescaled between 0 and 1 and thus are independent from image size.}
   \label{fig:fresco-centrality}
\end{figure}

First, in visual semiotics the \textit{spatial disposition} of each element can be determined with respect to the image frame, often represented in terms of oppositions (\textit{central} vs. \textit{peripheral} 1.3.1, \textit{left} vs. \textit{right} 1.3.2, \textit{high} vs. \textit{low} 1.3.3, \textit{foreground }vs. \textit{background} 1.3.4). In FRESCO.v1, we compute the position of the centroid of each identified object or person with respect to the vertical and horizontal midlines, as well as the distance from the center of the image (centrality), as illustrated in Figure \ref{fig:fresco-centrality}. Since positions are rescaled between 0 and 1, a value greater than or lower than 0.5 distinguishes between upper/left/peripheral and lower/right/central. As an approximation of whether an object is in the foreground or background, we compute the average depth by using a combination of panoptic segmentation and depth estimation. In visual semiotics, the figurative and plastic levels establish a complex interplay. At the computational level, this can be made evident by determining how we chose to partition the image into its constituent forms and elements. In FRESCO.v1, which focuses mostly on photography, elements are defined at the figurative level, through object detection and panotic segmentation. Other forms or elements could be extracted purely on the basis of plastic or compositional features (e.g., texture segmentation). This aspect should be kept in mind to account for future extensions.

Then, the spatial disposition of the elements with respect to each other is determined (1.3.5). In CV terms, these spatial relationships can be interpreted as a special case of the more general task of visual relationship detection \citep{cheng2022visual}. Spatial coverage (that is, the percentage covered by each class in a semantic segmentation) is also an indirect indicator of the spatial arrangement of objects \citep{mannisto2022automatic}. 

Plastic analysis also deals with how different forms and compositional elements interact with each other and how these interactions can shape the viewer's interpretation. Meaning is evoked by forms by assigning them qualities, which derive both from their internal characteristics and, above all, from the network of \textit{spatial, temporal and cooperative or contrastive relationships} with other forms and the surrounding space \citep{eugeni2014analisi}. These connections can be established among \textit{adjacent} (1.4) or \textit{distant} (1.5) forms based on their \textit{similarities} (e.g., same shape) or \textit{differences} (e.g., dark vs. light). These connections are independent of the figurative content of the respective forms and may thus reinforce or redefine the interpretation that may be formed based on the figurative content alone. 

Finally, all plastic elements contribute to the overall configuration (1.6), which can be balanced (predominantly static) or unbalanced (predominantly dynamic) (1.6.1).

% secondo paragrafo: figurative
The \textit{figurative} level is concerned with the main topic (2.1), persons, objects, scene and setting (2.2), movement (2.3), actions (2.4) and emotions (2.5). 

The characterization of individual and groups of people is partially based on AICE \citep{mannisto2022automatic}, which in turn is based on the work of \citet{kress1996reading}. We do not distinguish explicitly between attributes of the main character and of the side character, but assume that the categorization is available for each character, and expand the characterization to include attributes available in pre-trained facial attribute extractors \citep{zheng2022general}. 

\citet{eugeni2014analisi} distinguishes among movement (2.3) and actions (2.4). The movement categories pertain to how the image, which by nature is static, captures the evolving temporal dynamics of the scene. Over time, different strategies have been evolved to suggest how the scene depicted articulates in time, so that the view can evoke the temporal continuity that the still image cannot physically represent. Such techniques can be differentiated based on whether the image represents one or more instants in time within the same frame, which are classified by \citet{eugeni2014analisi} in \textit{ blocked, contracted, or articulated} (2.3.1). The resulting configuration may shun realism in favor of making the articulation of movement \textit{manifest} in the image (2.3.2); otherwise, the articulation of movement is assumed to be \textit{hidden} in the presentation (2.3.2). While semiotics deals with all forms of still images, not only photographs but also paintings and illustrations, in FRESCO.v1 we concentrate on photographs and especially social image images, which are likely going to represent a single instant in time (blocked). 
Actions, on the other hand, refer to the semantic interpretation of the depicted gestures and interactions (2.4). \textit{Single actions} (2.4.1) can be associated with CV tasks such as pose estimation and action recognition. In the case of \textit{ aggregate of actions} (2.4.2) or narratives (2.4.3), estimating the scene graph or detecting visual relationships would be necessary to differentiate the gestures performed by different characters and capture interactions among them. 
Finally, at the figurative level, we measure the \textit{intensity} (2.5.1) and \textit{class} of the \textit{emotions} (2.5.2) expressed by the characters depicted in the image. Some works in the CV literature have also investigated how to determine emotional valence (2.5.3), that is, the emotion aroused by the image \citep{lu2016identifying} in the viewer.

To conclude the figurative level, we also included as part of FRESCO.v1 image tagging \citep{huang2023inject} and visual captions \citep{hu2022scaling}. These models have the advantage of being trained on extremely large-scale datasets and thus were designed to achieve strong open-set capabilities, which are essential in the context of social media. At the same time, textual descriptions cannot be easily mapped to a specific figurative element, and thus may pose some issues when interpreting the results. 

At the \textit{enunciational level}, we focus in particular on the construction of the point of view of the \textit{spectator} (\textit{basic watcher-looked system} 3.1) and the \textit{observer} (\textit{first grade secondary watcher-looked system} 3.2). The former, in photography, essentially coincides with the position of the camera. It can be reconstructed on the basis of aspects such as perspective (horizon line and vanishing points), the position of the camera with respect to the scene, and the distance between the main subject(s) and the viewer. It is important to distinguish close-up and portraits from indoor and outdoor scenes, since the position of the spectator cannot always be clearly defined and is inferred based on different compositional cues depending on the type of image. 

The way a photograph is framed, and therefore what is not shown as well as what it is shown, is of paramount importance to shape its interpretation. By \textit{observer}, we denote a character that is explicitly never depicted, but implicitly assumed by a composition. The position of the observer can be inferred by the \textit{pose} of the characters (\textit{body pose} 3.2.2 and \textit{head pose} 3.2.1), and most importantly by the \textit{direction of the gaze} (3.2.3). The \textit{relative position} of the spectator and observer (3.4.1) will elicit involvement or detachment in the viewer, depending on whether they coincide or differ.  The presence of bystanders, indicators, insignias, and epigraphs \footnote{In semiotics, bystanders and indicators refer to secondary characters that are looking, pointing or otherwise directing the viewers' attention to the main focus of the scene. Insignias and epigraphs are objects (such as mirrors) or spatial elements of the scene (such as the presence of doors, or the direction of the light) with similar function. } further helps to guide the viewer in the correct interpretation, with a higher level of guidance reflecting in a greater sense of participation, especially in artistic composition \citep{eugeni2014analisi}.

Several variables evaluated in FRESCO involve estimating the distances between the observer (the camera, in the case of a photographic image) and the subject(s) depicted, as well as among the subjects depicted in the images in the case of multiple subjects. We estimate the distance of the main character combining the depth map and the panoptic segmentation considering only the \quoten{person} category. Similarly, we compute also the distance of main subject taking into account all the \quoten{things} categories, which includes only countable objects, since the primary theme in a photo may not necessarily be a person. Interpersonal distances are shaped by our sensory-motor possibilities (e.g., whether we can touch, hear, or smell another person) but are also influenced by social and cultural conventions; hence, they carry with them a plethora of implicit messages. Moving from the seminal works of \citet{hall1966hidden}, one of the cardinal findings of proxemics dictates that people tend to organize the space around them in terms of four concentric zones (intimate zone, casual personal zone, social zone, and public zone) associated with increasing degrees of intimacy and interactions. This classification forms the basis for subsequent works in visual semiotics \citep{bell2012content}, as well as in computational visual proxemics or Visual Social Distancing (VSD) estimation, that is, approaches that rely on cameras and other imaging sensors  to analyze the proxemic behavior of people \citep{cristani2011towards,cristani2020visual}.

\begin{table*}[!tbh]
\centering
\resizebox{\linewidth}{!}{%
\begin{tabular}{ccccc}
\hline
\textbf{Task} & \textbf{Model} & \textbf{Dataset (train)} & \textbf{Dataset (test)} & \textbf{Performance (expected)} \\ \hline
Face Detection & RetinaFace (ResNet50) \citep{deng2020retinaface} & WIDERFACE (train) & WIDERFACE (val) & mAP: 96.5\%(easy), 95.6\% (medium), 90.4\% (hard) \\ \hline
Face Mesh & MediaPipe \citep{lugaresi2019mediapipe} & Private & Private & IOD MAD: 3.96\% \\ \hline
Head Pose & 6DRepNet \citep{hempel20226d} & 300W-LP & AFLW2000 & Yaw: 3.63, Pitch: 4.91, Roll: 3.37, MAE: 3.97 \\ \hline
Gaze Direction & 3DGazeNet \citep{ververas20223dgazenet} & Gaze360 (train) & Gaze360 (test) & Gaze error (degrees): 9.6 \\ \hline
Emotion Estimation & EmoNet \citep{toisoul2021estimation} & AffectNet (train) & AffectNet (test) & Expression Acc: 0.75 \\
 &  &  &  & Valence CCC: 0.82, PCC: 0.82, RMSE: 0.29, SAGR: 0.84 \\
 &  &  &  & Arousal CCC: 0.75, PCC: 0.75, RMSE: 0.27, SAGR: 0.80 \\ \hline
Face Attribute Estimation & FACER \citep{zheng2022general} & LAION-Face-20M & CelebA (test) & Acc: 92.1\% \\
 &  & + CelebA (train) &  &  \\ \hline
Age Estimation & DeepFace \citep{serengil2021hyperextended} & IMDB-WIKI & IMDB-WIKI & MAE: 4.65 \\ \hline
Gender Estimation & DeepFace \citep{serengil2021hyperextended} & IMDB-WIKI & IMDB-WIKI & Acc: 97.44\%, Precision: 96.29\%, Recall: 95.05\% \\ \hline
Ethnicity Estimation & DeepFace \citep{serengil2021hyperextended} & FairFace (train) & FairFace (test) & Acc: 68.0\% \\ \hline
Image Tags & RAM++ (Swin-L) \citep{huang2023inject} & COCO & OpenImages, & Tag-Common mAP: 86.6 (OpenImages), 72.4 (ImageNet-Multi)\\
& & + VG & ImageNet-Multi, & Tag-Uncommon mAP: 75.4 (OpenImages), 55.0 (ImageNet-Multi)\\ 
& & + SBU captions & HICO & Phrase-HOI mAP: 37.7 (HICO)\\ 
& & + Conceptual Captions & & \\ 
& & + Conceptual 12M & & \\ \hline
Scene Classification & VGG-Places365 \citep{zhou2017places} & Places365 (train) & Places365 (test) & Top-1 acc: 55.19\%, Top-5 acc: 85.01\% \\ \hline
Body Pose & PifPaf \citep{kreiss2019pifpaf} & COCO keypoint & COCO keypoint (test-dev) & AP: 66.7, AP$^{M}$: 62.4, AP$^{L}$: 72.9 \\ \hline
Depth Estimation & DPT-Hybrid \citep{ranftl2021vision} & MIX-6 & DIW & WHDR: 11.06 \\ \hline
Surface Normal & NLL-AngMF \citep{Bae2021} & ScanNet & ScanNet (test) & Angular error (degrees) Mean: 11.8, Median: 5.7, RMSE: 20.0 \\ \hline
Edge Detection & DexiNed-a \citep{poma2020dense} & BIPED & BIPED (test) & ODS: 0.859, OIS: 0.867, AP: 0.905 \\ \hline
Object Detection & UniDet \citep{zhou2022simple} & COCO & COCO (test), & mAP: 52.9 (COCO), 60.6 (OpenImages), \\ 
& & + Objects365 & OpenImages (test), & 25.3 (Mapillary), 33.7 (Objects365) \\
& & + OpenImages & Mapillary (test), & \\
& & + Mapillary & Objects365 (valid) & \\ \hline
OCR & CharNet \citep{xing2019charnet} & SynthM & ICDAR 2015 & Acc: 71.6 (sen), 74.2 (in-sen) \\ \hline
Semantic Segmentation & Mask2Former (Swin-L) \citep{cheng2022masked} & ADE20k & ADE20K (val) & mIoU (s.s.): 56.1, mIoU (m.s.): 57.3 \\ \hline
Panoptic Segmentation & Mask2Former (Swin-L) \citep{cheng2022masked} & COCO panoptic (train2017) & COCO panoptic (val2017) & PQ: 57.8, PQ$^{th}$: 64.2, PQ$^{st}$: 48.1, AP$^{th}_{pan}$: 48.6, mIoU$_{pan}$: 67.4 \\ \hline
Caption Generation & Prismer$_{LARGE}$ \citep{liu2023prismer} & Pre-train: COCO Caption (Karpathy train) & COCO Caption (Karpathy test) & BLEU@4: 40.4, METEOR: 31.4, CIDEr: 136.5, SPICE: 24.4 \\ 
& & + Visual Genome & & \\
& & + Conceptual Captions & & \\
& & + SBU captions & & \\
& & + Conceptual 12M & & \\
& & Fine-tune: COCO Caption (Karpathy train) & & \\ \hline
\end{tabular}%
}
\caption{Models included in FRESCO v1 implementation.}
\label{tab:implementation}
\end{table*}

\color{black}

\subsection{Implementation}
\label{ssec:implementation}
FRESCO relies on a collection of open-source, off-the-shelf CV models representing the state of the art in their respective tasks. Although we recognize that potentially more accurate results could be achieved using cloud-based commercial APIs, for the sake of privacy, reproducibility, and transparency, open implementation was preferred \citep{santangelo2023}.

\begin{figure*}
    \includegraphics[width=\textwidth]{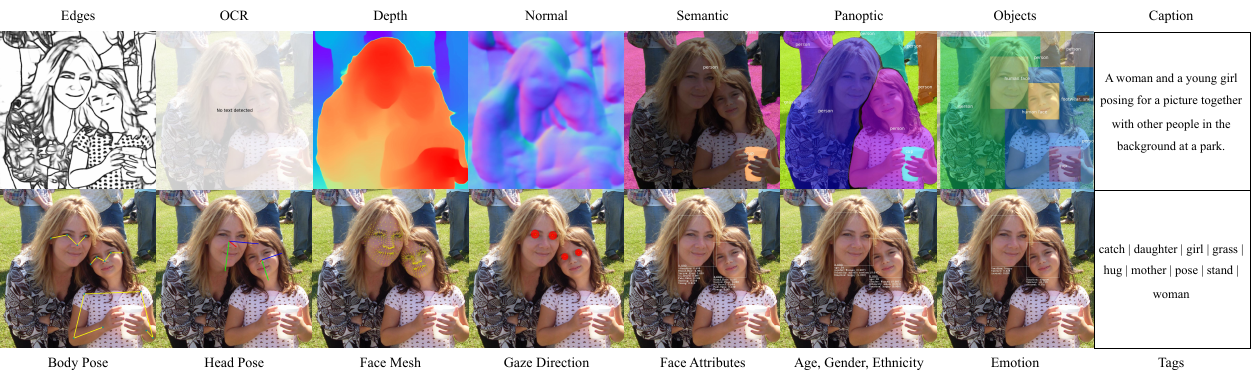}
    \caption{Example of output of models included in FRESCO.v1 implementation. }
    \label{fig:fresco-intermediate-output}
\end{figure*}

\paragraph{Built-in models} 
FRESCO.v1 includes the following models, whose key details are summarized in Table \ref{tab:implementation}. Face detection (1) is obtained using RetinaFace \citep{deng2020retinaface} with a ResNet50 backbone. The face mesh (2) is obtained from MediaPipe \citep{lugaresi2019mediapipe}, while the body pose (3) is obtained using PifPaf \citep{kreiss2019pifpaf}. Head pose (4) is estimated from 6DRepNet \citep{hempel20226d}, while gaze direction (5) is extracted using 3DGazeNet \citep{ververas20223dgazenet} using the InsightFace implementation. Continuous levels of valence/arousal (6) and emotion category (6) are estimated using EmoNet \citep{toisoul2021estimation}, while 40 facial attributes (7), corresponding to those available in the CelebA dataset \citep{liu2015faceattributes}, are extracted using FACER \citep{zheng2022general}. Age (8), gender (9), and ethnicity (10) are estimated using the DeepFace \citep{serengil2021hyperextended} framework. Depth estimation (11), edge detection (12), object detection (13), OCR (14), semantic segmentation (15), panoptic segmentation (16) and caption generation (17) are derived through PRISMER \citep{liu2023prismer} and its associated expert models \citep{ranftl2021vision,poma2020dense,zhou2022simple,liu2018char,cheng2022masked}. Image tags (18) are extracted using the Recognize Anything Model - RAM++ \citep{huang2023inject}. Scene classification (19) is performed using a VGG model trained on Places365 \citep{zhou2017places}. Chromatic information (20) is extracted using established image processing techniques, while simple geometric measures are implemented custom. An example of the output of these models can be seen in Figure \ref{fig:fresco-intermediate-output}.

\paragraph{Structured data extraction} 
We tightly combined the output of these pre-trained models with geometric properties and/or image processing methods to extract the information described in Section \ref{sec:fresco-design} and which correspond to the items marked by \checkmark in Table \ref{tab:fresco}. 

At the \textit{Plastic level}, the \textit{Eidetic features}, which in the current version includes only the line map (1.1.6), are obtained from the edge detector (12). \textit{Chromatic features} (1.2.1-3) are extracted using image processing techniques (20). For the \textit{Topological features}, those related to spatial disposition (1.3.1-3) are obtained from a direct comparison of the centroids' positions, derived from the bounding boxes found by the object detector (13), and the image area, resulting in three different positional ratios ranging in [0,1]. These values can be discretized to obtain the position of each object as Top/Center/Bottom, Left/Center/Right and Central/Periperal as exemplified in Figure \ref{fig:fresco-centrality}. In the current implementation, each centroid position is discretized considering the image area divided in three bands in the proportion 40:20:40, both vertically and horizontally, for the first two measures. An object is instead considered central if its centroid falls within an ellipse having semiaxes equal to 60\% of the image semiaxes. All these thresholds were chosen empirically and can be adjusted through parameters. The depth positions (1.3.4) are obtained by combining the output of the panoptic model (16) and the depth estimator (11). Specifically, the depth maps of each person and object detected in the image are isolated by masking the whole depth map with each instance segmentation map found by the panoptic and then averaged to obtain the average depth for each instance. The background average depth is obtained by averaging the remaining part of the depth map once all pixels corresponding to objects and people are removed. For spatial disposition (1.3.5), we estimate spatial coverage by computing the percentage of pixels covered by each class on the semantic segmentation map (15). 

At the \textit{Figurative level}, the main topic (2.1.1) is estimated using the image tagging model (18). The number of people (2.2.1.1) is derived from the output of the object detector (13). Even though the face detector (1) is highly accurate in its task, we decided to rely on the object detector to also consider people with occluded faces or photographed from behind. The characteristics of each person (2.2.2) are obtained by running on each face crop extracted by the face detector (1) the respective model for each task including the age (8), gender (9), ethnicity (10), face attributes (6). The object category and the text in the image (2.2.3.2-3) are obtained using the labels found by the object detector (13) and the OCR model (14). Scene characteristics (2.2.4.1 and 2.2.4.3-4) are inferred from the output of the scene classification model (19). For action (2.4.1), the body pose and the image caption are extracted using (3) and (17). Similarly to the single person characteristics, the emotions (2.5.1-3) measures are evaluated on each face crop extracted by the face detector (1) and using the emotion estimator (6) that returns both continuous values (valence/arousal) and emotion category. 

At the \textit{Enunciational level}, for the \textit{Basic watcher-looked system}, the distances from the camera (3.1.1-2) are estimated following an approach similar to the one adopted for (1.3.4). The framing (3.1.3) is instead obtained by computing the ratio between the area of the largest face crop found by the face detector (1) and the entire area of the image. In the current implementation, we consider an image as a portrait if the face crop covers more than 30\% of the total image or as a scene otherwise; the threshold can also be adjusted through a parameter. For the \textit{First-grade secondary watcher-looked system}, all angles (3.2.1, 3.2.3) related to head pose (yaw, pitch, roll) and gaze (yaw, pitch) are estimated on each face crop extracted by the face detector (1), by running the models for the head pose estimation (4) and gaze direction (5).

As a result, we obtain two sets of image-level functions $\mathcal{F}_t := \{f^k_t(\cdot)\}$, with $k=1 ... K$, and object-level functions $\mathcal{G}_t := \{g^l_t(\cdot)\}$, with $l=1 ... L$, and $t \in \{\texttt{plastic}, \texttt{figurative}, \texttt{enunciational}\}$, that take as input the whole image or at each single object detected in the image, respectively, for each level of analysis $t$, and that can be exploited to compare the content of the image as discussed in the following section. We make the implementation available at \url{https://gitlab.com/grains2/fresco}.

\color{black}

\section{The FRESCO Similarity Score}
\label{sec:fresco-score}
In this Section, we define FRESCO-Score, a similarity measure that leverages FRESCO, and specifically all the measures available in FRESCO.v1 as highlighted in Table \ref{tab:fresco}. It represents an estimation of how closely two images represent the same content at the plastic, figurative, and enunciational levels. Unlike feature-based similarity metrics \citep{ramtoula2023visualdnarepresenting,hessel2021clipscore,heusel2018gans}, FRESCO-Score allows for an in-depth exploration of which aspects of the images are mostly different and hence affect the final score offering significant benefits in terms of interpretability. In fact, two pairs of images may have comparable final distances, but differ with respect to heterogeneous aspects. For instance, a pair of images, despite having different chromatic properties, may depict a similar figurative content, and thus present a distance akin to another pair that, while similar at the plastic level, have different figurative contents. A user may also choose to weight each component differently, so as to cluster images based on specific properties.  These considerations are in general not possible with methods that return a single distance value between the representations of the two compared images in the feature space of a black-box pre-trained neural network. By considering all the analysis one by one, FRESCO-Score enables us to appreciate the difference among two images at different scales, delving into even the most intricate details, such as the direction of the gaze of each single person depicted in the image.

FRESCO-Score needs to aggregate properties that are associated with the whole image (e.g., chromatic categories, main topics, place) and single subjects or objects in the images (e.g., the characteristics, emotions, pose, and gaze of a specific person). While the former allows for a direct comparison of values computed at the image level, the latter requires a mapping strategy to associate each person/object of the first image to a comparable instance, if any, in the second one. For convenience, in the following we will refer to them as \emph{image-level} and \emph{object-level} measures. We exclude from FRESCO-Score intermediate maps (e.g., semantic maps or line maps) and body poses, that will be included in further development.

\color{black}
Specifically, given two images $I_i$ and $I_j$, FRESCO extracts a series of measurements using the set of image-level and object-level functions $\mathcal{F}_t := \{f^k_t(\cdot)\}$, with $k=1 ... K$, and object-level functions $\mathcal{G}_t := \{g^l_t(\cdot)\}$, with $l=1 ... L$, and $t \in \{\texttt{plastic}, \texttt{figurative}, \texttt{enunciational}\}$, defined in Section \ref{ssec:implementation}. Each image is associated with the set of objects $\{o_i^m\}_{m=0}^M$ and $\{o_j^n\}_{n=0}^N$ detected in each image and whose positions in the pixel area are described by the respective centroids $\{c_i^m\}_{m=0}^M$ and $\{c_j^n\}_{n=0}^N$.
FRESCO-Score first computes a matching function $m(\cdot, \cdot)$ that, given the two sets of centroids $\{c_i^m\}$ and $\{c_j^n\}$, returns a set of matched pairs $\{(\hat{c}_i^p, \hat{c}_j^p)\}_{p=0}^{P}$ such that a matching cost function is minimized.
Then, FRESCO-Score is computed as: 

\begin{equation}
    S = \alpha S_{\texttt{pla}}(I_i, I_j) + \beta S_{\texttt{fig}}(I_i, I_j) + \gamma S_{\texttt{enu}}(I_i, I_j)
\end{equation}
where $\alpha$, $\beta$, $\gamma$ are configurable parameters (set to 1 in the rest of this paper), and each $S_t$, with $t \in \{\texttt{plastic}, \texttt{figurative}, \texttt{enunciational}\}$, is computed by aggregating the pertinent subset of features as follows:

\begin{equation}
    %S_t(I_i, I_j) = \delta^k\sum_{k=1}^K{d^k(f_t^k(I_i), f_t^k(I_j))} + \eta^l\sum_{l=1}^L{\epsilon^m\sum_{m=1}^M{d^l(g_t^l(\hat{o}_i^m), g_t^l(\hat{o}_j^m))}}
    S_t(I_i, I_j) = \sum_{k=1}^K{d^k(f_t^k(I_i), f_t^k(I_j))} + \sum_{l=1}^L{\sum_{p=0}^P{d^l(g_t^l(\hat{o}_i^p), g_t^l(\hat{o}_j^p))}}
\end{equation}
where $(\hat{o}_i^p, \hat{o}_j^p)$ are the pairs of matched objects, $\mathcal{F}_t := \{f^k_t(\cdot)\}$ and  $\mathcal{G}_t := \{g^l_t(\cdot)\}$ are the subset of functions used to extract the measures at each level $t$, and $d^k$ and $d^l$ are the normalized distance functions computed on each couple of image- and instance-level measures, respectively. Prior to aggregating, distances can be scaled in the range [0, 1], as done in the remainder of this paper, or standardized using a mean and a standard deviation precomputed on a large dataset.

\color{black}

\paragraph{Image-level measures} Palette similarity is derived from the CIELAB color difference between two single colors obtained as the weighted average of the two palettes, following the single (homogeneous) color difference model proposed in \citet{pan2018comparative}. 

\color{black}
RGB color histograms are compared using the Hellinger distance, which is related to the Bhattacharyya coefficient as follows:
\begin{equation}
    BC(H_i, H_j) = \sum_{x \in \mathcal{X}}\sqrt{H_i(x) \cdot H_j(x)}
\end{equation}
\begin{equation}
    d(H_i, H_j) = \sqrt{1 - BC(H_i,H_j)}
\end{equation}
where $H_i$ and $H_j$ are the histograms of the two images $I_i$ and $I_j$. The final distance is obtained as the average of the distances computed across three channels.
\color{black}
For scalar measures such as brightness, saturation, face-background ratio and background average depth, the absolute error is considered. Binary measures such as grayscale and indoor/outdoor evaluate to 1 if the corresponding value is the same in both images, 0 otherwise. Scene classification is compared using the cosine similarity on the confidence vectors returned by the model. Image tags are compared using the Jaccard index \citep{real1996probabilistic}, while for the spatial coverage a continuous Jaccard index was properly designed, taking into account the common area for each category. The number of people and objects in the image are compared using the percentage of common instances. The caption is instead compared using the cosine similarity between the text embeddings extracted by the CLIP ViT / L-14 text encoder. All distances are scaled in the range [0,1].

\paragraph{Mapping strategy} In FRESCO.v1 each instance in the first image is associated with the closest instance of the same category in the second image using the centroids derived from the bounding boxes (faces and objects) or the instance masks (for analysis involving depth information). 
 
\color{black}
The set of centroids $C_i := \{c_i^m\}_{m=0}^M$  identified in the first image $I_i$ is associated with the set of centroids $C_j := \{c_j^n\}_{n=0}^N$  found in the second image $I_j$, minimizing the cost of matching. Specifically, given the two sets $C_i$ and $C_j$, and a matching cost function $E: C_i \times C_j \rightarrow \mathbb{R}$, the objective is to find a bijection $f: C_i \rightarrow C_j$ such that the total cost of matching $\sum_{c_i\in C_i}E(c_i, f(c_i))$ is minimized.
The cost $E$ is defined as the squared Euclidean distance between each pair of centroids in the bipartite graph.
\color{black}
To solve this problem, we leverage the SciPy’s modified Jonker-Volgenant algorithm  for linear sum assignment, which has a complexity of $O(n^3)$ in the worst case \citep{crouse2016implementing}.

\paragraph{Instance-level measures} All instances in the two images are compared one-by-one after executing the mapping algorithm. Object positions (vertical ratios, horizontal ratios, centralities, distances from cameras) are compared using the absolute error. Person characteristics are compared using the cosine similarity on the confidence vectors returned by the models for both multiattribute (i.e., 40 face attributes and gender) and multiclass (i.e., ethnicity and emotion) classifications. Continuous values such as age, valence, arousal and angles (roll, pitch, yaw) for both head pose and gaze direction are compared using the absolute error. Unpaired objects and faces (i.e., those for which the matching algorithm does not return a real association) are assigned by default to the minimum similarity value for all the associated measures. All distances are defined in or are scaled to the range [0,1].

\begin{figure}[tb]
    \centering
    \includegraphics[width=\linewidth]{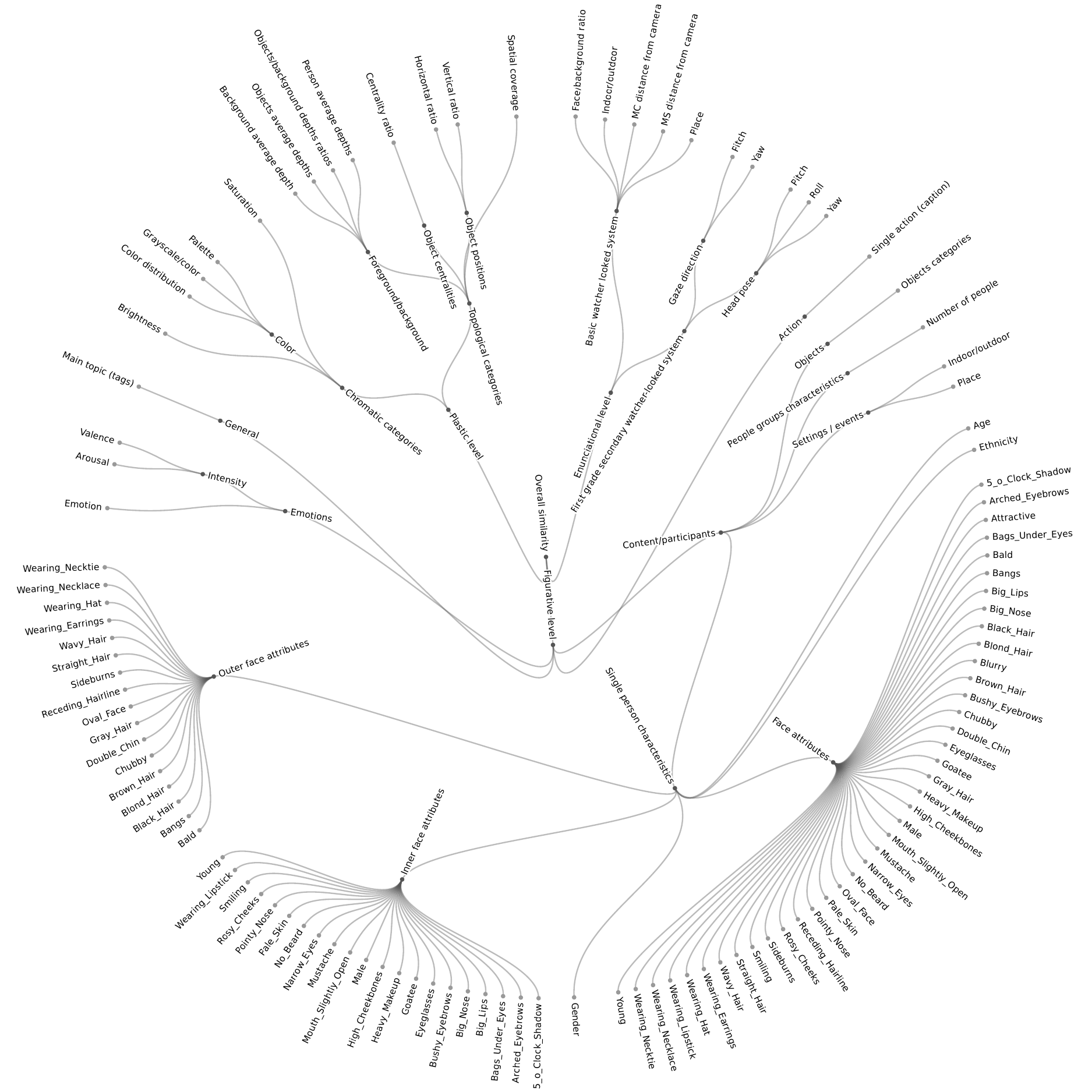}
    \caption{The hierarchy of similarities on which Fresco Score is built. Best viewed online. }
    \label{fig:aggregations}
\end{figure}

\paragraph{Aggregation} In FRESCO.v1, all similarity values are aggregated using a linear weighted average. Instance-level similarities are combined to obtain an image-level value for each analysis. Specifically, the similarities computed for each characteristic of each individual face in the image are averaged to derive a single compound measurement. Likewise, similarities related to single objects were also averaged out. Similarity scores are further aggregated to obtain a cumulative measure for groups of correlated characteristics (e.g., chromatic, topological, etc.), and then each group of measures is further averaged to get a per-level measure. The final overall score is obtained by averaging the measures of the three levels of analysis (i.e., plastic, figurative, and enunciational). 
An overview of the hierarchy of similarity scores and their relative aggregations is illustrated in Figure \ref{fig:aggregations}. It should be noted that, while all individual similarities are scaled between 0 and 1, their distribution may differ in practice. In the future, a more sophisticated aggregation will be considered, in which distance measures are calibrated on a target population. 

\section{Experimental validation}
\label{sec:experimental}
\subsection{Datasets}
Experimental validation was carried out on the FFHQ in-the-wild validation set \citep{karras2017progressive} composed of 10,000 images and on the MIAP (More Inclusive Annotations for People) \citep{miap_aies} extension of the OpenImages v7 database \citep{benenson2022colouring,kuznetsova2020open}, which we further filtered by specifically considering the presence of at least one visible human face, obtaining a total of 2002 images. The former dataset is composed of uncropped original images scraped from Flickr. It encompasses a wide range of variations in terms of age, ethnicity, and image background. In addition, it also features a relevant assortment of accessories, including eyeglasses, hats, and more. The latter was instead selected because of its greater variability of contents, which includes more complex scenes. It is also equipped with additional annotations with respect to the original OpenImages including exhaustive bounding boxes for all people and attribute labels such as the human perceived gender and perceived age range.
\subsection{Validation}
There are two fundamental questions that must be kept in mind when validating a computational pipeline such as FRESCO, and concern its validity (i.e., how faithful is the information extracted from FRESCO to the original images) and usefulness (i.e., to what extent the measures extracted by FRESCO can be used to answer interesting research questions that complement and extend traditional manual extraction). 

In this paper, we focus first and foremost on assessing the validity of the information extracted. Although each component included in FRESCO.v1 has been previously tested in isolation (relevant performance metrics are included in Table \ref{tab:implementation}), some residual errors (both random and systematic) are unavoidable. These aspects are here investigated by searching for discrepancies and inconsistencies in the output of different tools, which not only provides an indirect measure of performance, but can also guide researchers from different fields in interpreting the results.  The FRESCO-Score was validated by visually analyzing how it ranked images in terms of similarities.

\section{Results}
\label{sec:results}
\begin{figure*}[!tbh]
    \centering
    \includegraphics[width=\textwidth]{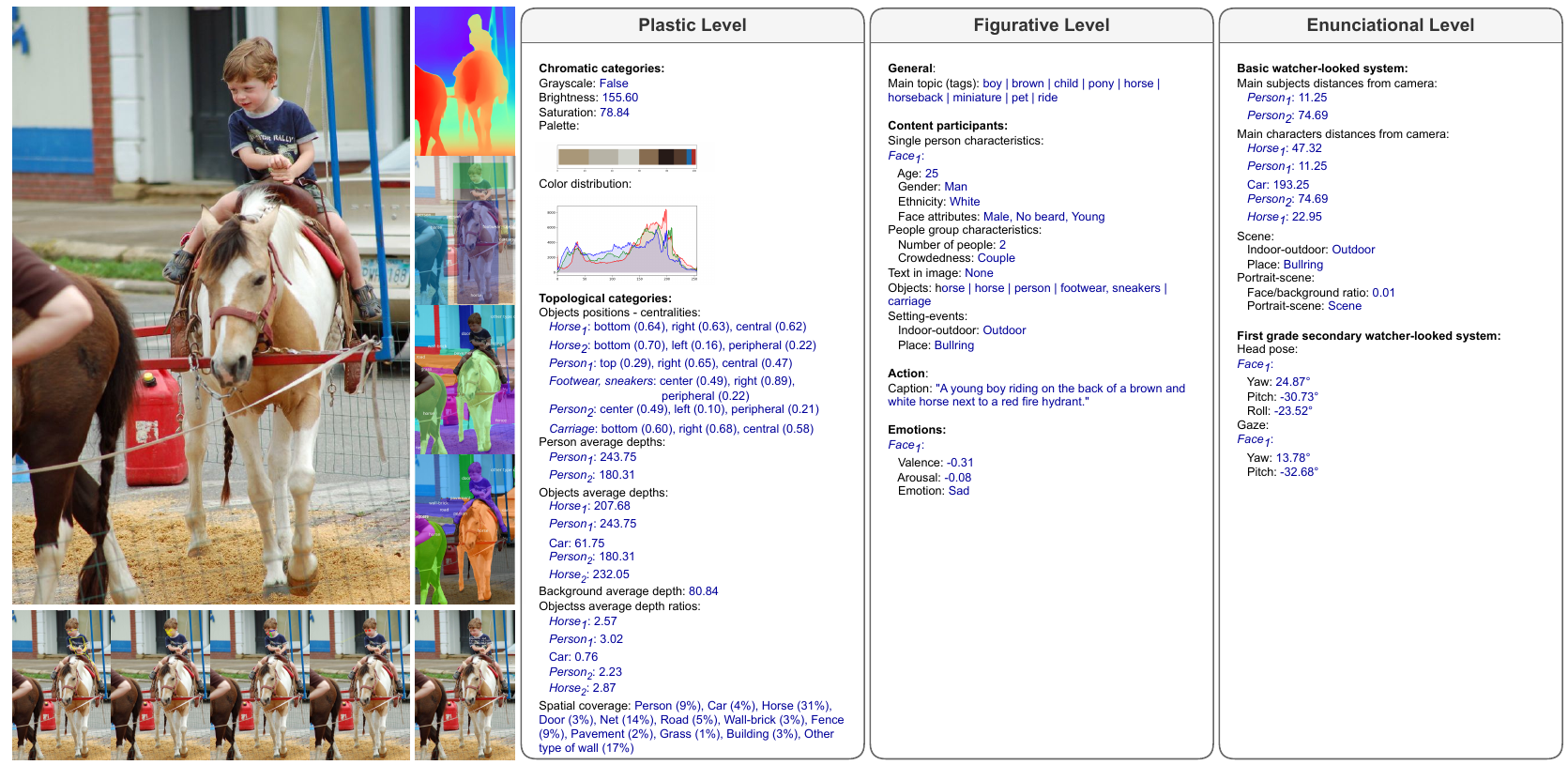}
    \caption{Example of FRESCO.v1 final output. All quantities extracted are defined in Table \ref{tab:fresco}. The figurative and plastic level are closely intertwined. Notice, for instance, how the figurative content is that of a boy riding a horse, but the spatial disposition of the figures is part of the plastic level (the horse is positioned in the center and occupies a substantial portion of figure,  and that the boy is located in the top right part of the image).  }
    \label{fig:fresco-final-output}
\end{figure*}

An example of application of the FRESCO.v1 computational pipeline is depicted in Figure \ref{fig:fresco-final-output}. It should be noted that many of the CV tools included in FRESCO.v1 produce a spatial output, such as a depth map or segmentation map, as exemplified in Figure \ref{fig:fresco-intermediate-output} and in the left part of Figure \ref{fig:fresco-final-output}. Keypoints and spatial maps are unstructured data and are difficult to analyze at scale; for this reason, FRESCO.v1 converts them into a set of interpretable indicators, either numerical or categorical, that lend themselves to analytics techniques. An example is provided in the right part of Figure \ref{fig:fresco-final-output}, in which quantities are divided according to whether they are pertinent to the plastic, figurative, or enunciational level. 

In the rest of this Section, we first evaluate the consistency of models (Section \ref{ssec:results_consistency}) and then proceed to evaluate the FRESCO-score similarity measure (Section \ref{ssec:results-score}).

\subsection{Accuracy and consistency of the extracted quantities}
\label{ssec:results_consistency}

First, we seek to answer questions related to how the output of different measures can be combined and compared. For instance, in Figure \ref{fig:fresco-final-output} it is evident that the estimated age is incorrect (25 years). However, both the caption and the tags refer to the presence of a young boy or a child, which is an indication that the age may not be accurately estimated. Two persons and two horses are detected by semantic segmentation and object detection, but only one pose and face are detected: indeed, an arm and the back of a horse are visible in the bottom left corner. An uncommon object (a red tank) is misclassified by the caption as a fire idrant, but the latter is not identified by the object detector, nor the semantic segmentation. A more systematic analysis of these discrepancies is presented in Tables \ref{tab:fresco_output_consistency_people} and \ref{tab:fresco_output_consistency_topics}. 

\begin{table}[tb]
\centering
\resizebox{\linewidth}{!}{
\begin{tabular}{c|cc}

& \multicolumn{2}{c}{\textbf{People detected (OpenImages)}} \\ 
\hline

\hline

\textbf{Tasks} & Images with people & People per image \\
& (at least one) & (all detected)\\ \hline 

\hline
Face detection & 90.26\% & 2.07 \\ \hline
Object detection & 88.31\% & 1.85 \\ \hline
Panoptic segmentation & 92.01\% & 2.14 \\ \hline
Semantic segmentation & 95.55\% & - \\ \hline
Tagging & 96.10\% & - \\ \hline
Captioning & 95.35\% & - \\ \hline

\\

& \multicolumn{2}{c}{\textbf{People detected (FFHQ in-the-wild)}} \\ 
\hline

\hline

\textbf{Tasks} & Images with people & People per image \\
& (at least one) & (all detected) \\ \hline 

\hline
Face detection & 100.00\% & 1.75 \\ \hline
Object detection & 99.21\% & 2.03 \\ \hline
Panoptic segmentation & 99.84\% & 2.27 \\ \hline
Semantic segmentation & 99.97\% & - \\ \hline
Tagging & 99.99\% & - \\ \hline
Captioning & 99.98\% & - \\ \hline

\hline
\end{tabular}%
}
\caption{Analysis of the people detection consistency across different models included in FRESCO.v1 evaluated on OpenImages and FFHQ in-the-wild datasets.}
\label{tab:fresco_output_consistency_people}
\end{table}

\begin{table*}[!tbh]
\centering
\resizebox{0.75\textwidth}{!}{
\begin{tabular}{c|cccccc}

& \multicolumn{6}{c}{\textbf{Images with groups of people (OpenImages)}} \\ 
\hline

\hline

\textbf{Tasks} & 0/1 & 2 & 3-6 & 7-12 & 13-30 & 31+ \\
& (single) & (couple) & (small group) & (medium group) & (large group) & (crowd) \\ \hline 

\hline
Face detection & 67.88\% & 15.78\% & 11.94\% & 2.85\% & 1.25\% & 0.30\% \\ \hline
Object detection & 60.84\% & 17.38\% & 18.63\% & 3.00\% & 0.15\% & 0.00\% \\ \hline
Panoptic segmentation & 61.04\% & 16.28\% & 16.18\% & 5.54\% & 0.95\% & 0.00\% \\ \hline

\\

& \multicolumn{6}{c}{\textbf{Images with groups of people (FFHQ in-the-wild)}} \\ 
\hline

\hline

\textbf{Tasks} & 0/1 & 2 & 3-6 & 7-12 & 13-30 & 31+ \\
& (single) & (couple) & (small group) & (medium group) & (large group) & (crowd) \\ \hline 

\hline
Face detection & 61.73\% & 21.80\% & 15.06\% & 1.17\% & 0.24\% & 0.00\% \\ \hline
Object detection & 52.70\% & 21.80\% & 23.56\% & 1.94\% & 0.00\% & 0.00\% \\ \hline
Panoptic segmentation & 51.99\% & 21.79\% & 20.60\% & 5.40\% & 0.22\% & 0.00\% \\ \hline

\hline
\end{tabular}%
}
\caption{Analysis of the people groups detection consistency across different models included in FRESCO.v1 evaluated on OpenImages and FFHQ in-the-wild datasets. In FRESCO.v1 counts are inferred from the object detector to adjust for individuals seen from behind.}
\label{tab:fresco_output_consistency_people_groups}
\end{table*}

\begin{table*}[!htb]
\centering
\resizebox{\textwidth}{!}{
\begin{tabular}{c|ccc|ccc|ccc}

& \multicolumn{9}{c}{\textbf{Topics detected (OpenImages)}} \\ 
\hline

\hline

\textbf{Tasks} & In first & In common & In second & In first & In common & In second & In first & In common & In second \\
& & ($CLIP_{Score} \geq 0.80$) & & & ($CLIP_{Score} \geq 0.85$) & & & ($CLIP_{Score} \geq 0.90$) & \\ \hline 

\hline
Tags-Objects & 60.12\% & 29.86\% & 10.01\% & 71.91\% & 11.58\% & 16.52\% & 75.16\% & 5.47\% & 19.37\% \\ \hline
Tags-Semantic & 50.37\% & 30.09\% & 19.53\% & 64.40\% & 9.23\% & 26.37\% & 67.00\% & 4.63\% & 28.36\% \\ \hline
Tags-Panoptic (things) & 72.03\% & 25.70\% & 2.27\% & 86.22\% & 7.82\% & 5.96\% & 88.21\% & 3.48\% & 8.31\% \\ \hline
Objects-Semantic & 23.16\% & 26.30\% & 50.54\% & 30.29\% & 16.21\% & 53.49\% & 30.83\% & 15.21\% & 53.97\% \\ \hline
Objects-Panoptic (things) & 47.94\% & 43.38\% & 8.68\% & 62.09\% & 27.96\% & 9.95\% & 62.69\% & 26.87\% & 10.43\% \\ \hline
Semantic-Panoptic (things) & 65.49\% & 34.51\% & 0.00\% & 71.84\% & 28.16\% & 0.00\% & 72.27\% & 27.73\% & 0.00\% \\ \hline

\hline
\end{tabular}
}
\caption{Analysis of the topics detection consistency across different models included in FRESCO.v1 evaluated on OpenImages dataset. It considers three different thresholds on the labels encoding similarities to establish if a topic can be considered in common among the predictions of each pair of models.}
\label{tab:fresco_output_consistency_topics}
\end{table*}

\textit{Determining whether two outputs are consistent }is not straightforward as FRESCO involves multiple models whose set of labels may differ substantially, requiring some form of concept mapping. Some concepts may be expressed only by one label set, or may be expressed by different label sets at varying granularities. The problem is more evident when models with a predefined, closed-set output label space, such as object detectors, are compared with open-set models such as tagging or captioning models. As an example, the caption generator can produce outputs in the form \textit{\quoten{A baseball player catching a ball…}}, \textit{\quoten{A family posing for a picture…}}, \textit{\quoten{A mother and daughter pose for a picture…}} that suggest the presence of people in the image even if the concept “person” is not explicitly mentioned. 
On the other hand, the object detector used in Fresco.v1 does not consider label hierarchies \citep{zhou2022simple}, hence the class \quoten{person} is considered as an independent concept with respect to \quoten{man}, \quoten{woman}, \quoten{girl} and \quoten{boy} which are also included in the label space.

 Table \ref{tab:fresco_output_consistency_people} reports \textit{the number of people (2.2.1.1) detected by each model or task}. To achieve a more reliable estimate of the number of people detected by each model, a broad synset was properly selected to represent the concept “person” including all labels and expressions that can be traced back to the original wider concept. The results show that the number of detected faces is lower than the number of people found by the object detector and the panoptic segmentation, especially on FFHQ in-the-wild. This should not necessarily be interpreted as an indication of poorer performance of the face detector, but, more likely, this result may be attributable to the presence of people photographed from behind or with occluded/cut faces. Unexpectedly, the panoptic model seems to retrieve a slightly larger number of persons.  It should be noted that, unlike object detection, the panoptic label space and semantic segmentation include both \quoten{things} and \quoten{stuff} categories. Consequently, they are able to catch information about both countable objects which are characterized by a well-defined shape (things) and uncountable categories which are in general amorphous and belong predominantly to the context of the scene (stuff). Hence, this difference may be explained by the huge gap in the number of categories taken into account by the two models, only 133 (80 \quoten{things} + 53 \quoten{stuff}) compared to 722. 
 The percentage of images recognized as containing people is close to 100\% for all models on FFHQ in-the-wild. Instead, despite the OpenImages split in use being properly filtered in the presence of \quoten{Human face}, the percentage is in general lower. Unlike FFHQ in-the-wild, in OpenImages a limited number of sketches and cartoons are included, since they were annotated as containing \quoten{Human face}; however, models such as the face detector are trained on real faces and may fail in these different domains. The models used for tagging and captioning appear to be more robust also in these images.  Last, in Table \ref{tab:fresco_output_consistency_people_groups} we report the distribution of the number of people discretized according to the categories defined in Table \ref{tab:fresco} (that is, no people, single person, couple, small group, medium group, large group or crowd).

\textit{The consistency among the topics (2.1.1) and the objects (2.2.3) detected by each model} was further evaluated on the OpenImages validation set (Table \ref{tab:fresco_output_consistency_topics}).
To make semantic similar labels comparable, we leveraged a variant of the CLIP score which evaluates the cosine similarity between the text embeddings of the two labels extracted using the CLIP ViT-L/14 text encoder. To establish whether a concept is equally recognized by different models, we set a threshold on the similarity score. Some labels, despite referring to the same concept, may use different words and/or include more details. For example, labels \quoten{land vehicle}, \quoten{sport car}, and \quoten{sedan} are more specific cases of the general concept \quoten{car}; compared to label \quoten{car}, their CLIP score is 0.83. Depending on the threshold selected, we may consider semantically related information as equivalent or not. We compared each pair of tasks (e.g., image tagging vs. object detection, image tagging vs. panoptic segmentation) to determine whether on average each task provides more, equal, or less information than the other (i.e., whether the output contains the same concept, or whether certain concepts are present only in one of the outputs), at a given CLIP score threshold. 
The results of Table \ref{tab:fresco_output_consistency_topics} indicate that image tagging can associate the highest number of topics with a given image. Compared to image tagging, semantic segmentation can identify the highest number of additional topics (around 20\% at a threshold of 0.8), followed by the object detector with about 10\%. Semantic segmentation grasps more concepts w.r.t. to object detection as it embraces both labels from countable objects (things) and uncountable categories (stuff) which characterize mainly the background. This is further supported by the results achieved by the semantic segmentation, which adds more than 50\% topics to the object detector. In turn, the object detector is able to find much more topics compared to the panoptic segmentation (things) due to its higher label space (722 vs. 80). Panoptic segmentation cannot detect more topics than semantic segmentation, as it was trained on the same dataset and its label space is a subset of the latter. In this case, the topics in common are likely to be the things identified by both models. Lastly, the use of multiple models trained on different datasets introduces a relevant benefit: it allows to capture a wide range of information from the image compensating any oversight of each single model. In fact, common topics are less likely to arise from mispredictions of individual models. 

\begin{figure*}[!tbh]
   \centering
   \includegraphics[width=0.45\linewidth]{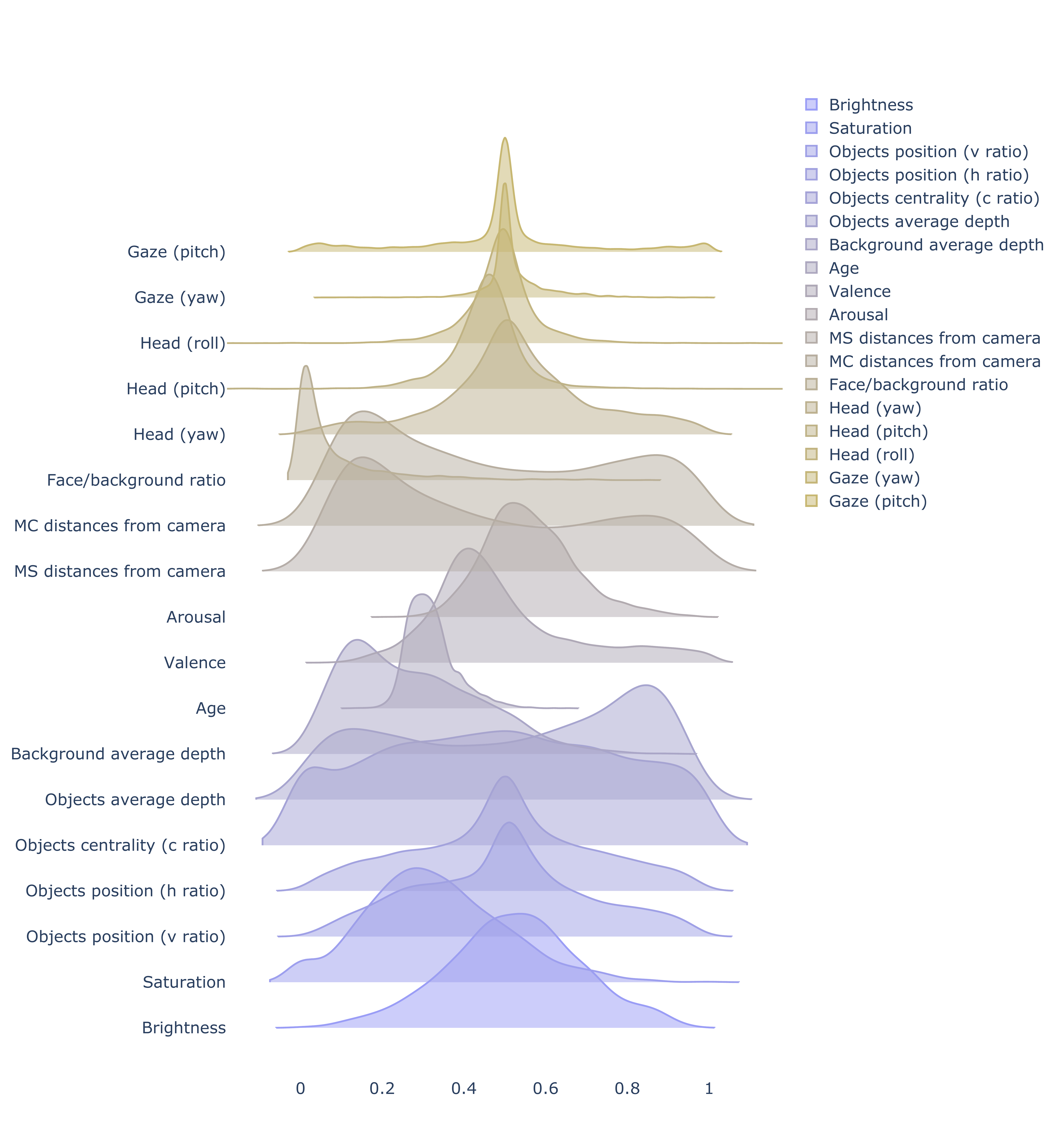}
   \includegraphics[width=0.45\linewidth]{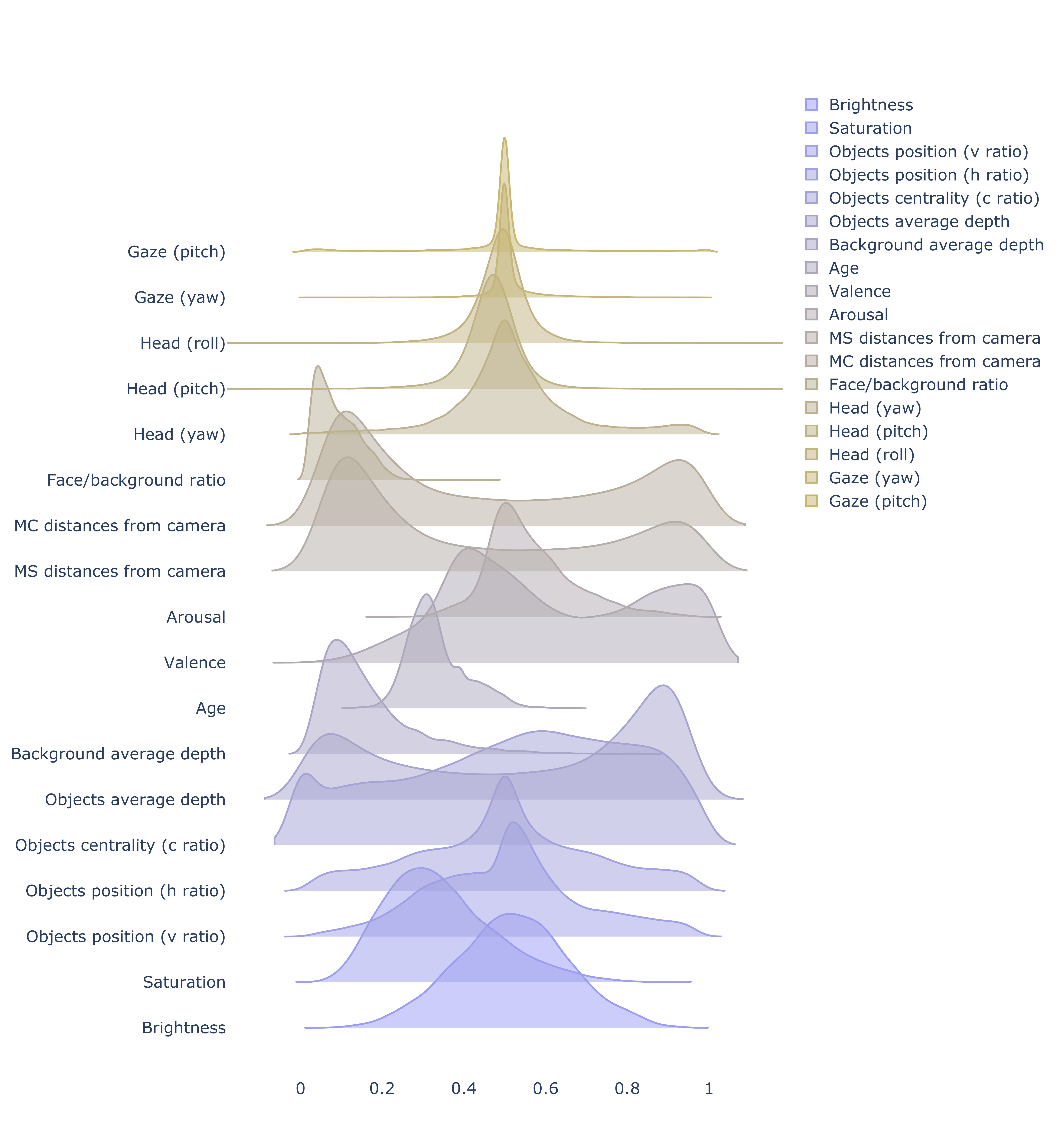}
   \caption{Distribution of a selection of numerical values in the OpenImages (left) and FFHQ-in-the-wild (right) subsets.}
   \label{fig:distributions}
\end{figure*}

\begin{figure*}[!tbh]
   \centering
   \includegraphics[width=0.9\textwidth]{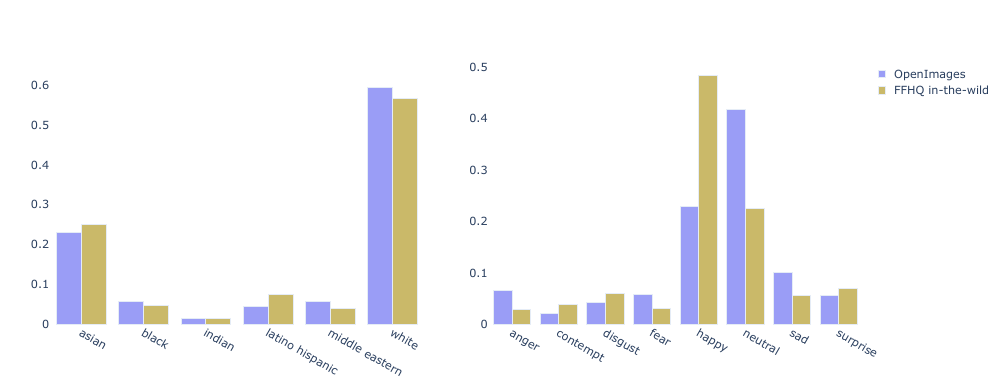}
   \caption{Distribution of a selection of categorical measures in the OpenImages and FFHQ-in-the-wild subsets.} 
   \label{fig:distributions-cat}
\end{figure*}

Finally, we investigated \textit{the distributions of a subset of continuous and categorical measures }extracted from the FFQH-in-the-wild and OpenImages validation set (Figures \ref{fig:distributions} and \ref{fig:distributions-cat}).  In terms of plastic categories, the centroids of objects and persons appeared to be predominantly located \textit{in the middle of} the picture frame both horizontally (1.3.2) and vertically (1.3.1). Both datasets have similar characteristics in terms of  \textit{brightness} (1.2.2) and \textit{saturation} (1.2.3). 

At the figurative level, all images depict close-up portraits or scenes in which two or more people interact. Unlikely OpenImages, FFHQ in-the-wild shows a bimodal distribution for the \textit{Valence} (2.5.3) category, which is consistent with a substantial presence of people smiling and posing for the camera. \textit{Emotion classification } (2.5.2) further supports this finding, since the FFQH in-the-wild distribution reaches its peak in the \quoten{happy} category, while for OpenImages the dominant class is \quoten{neutral}. The \textit{age} (2.2.2.2) distribution is quite similar for both datasets varying mostly in the range 20-50 (age is normalized between 0 and 100). In terms of \textit{ethnicity} (2.2.2.5), both datasets are imbalanced with a strong prevalence of \quoten{white} and \quoten{asian} categories, while others are markedly underrepresented. 

At the enunciational level, with a few exceptions (more evident in OpenImages), the gaze (3.2.3) and head (3.2.1) angles peaked around the value of 0.5, indicating the predominant presence of people looking at the camera while posing for pictures. The face/background ratio (3.1.3) is skewed in the 0-0.2 range, even more evident for the OpenImages split, suggesting that the majority of images are scenes depicting people in context, rather than portraits. The current implementation sets a threshold at 0.3, so an image is considered a portrait if the largest face box covers at least 30\% of the total image area. 

The \textit{main subject's distance from camera} (3.1.1) is a bimodal distribution, suggesting the presence of two main groups: close-up portraits and scenes in which several persons interact.

\subsection{Validation of the FRESCO score}
\label{ssec:results-score}

\begin{figure*}[!tbh]
    \centering
    \includegraphics[width=\textwidth]{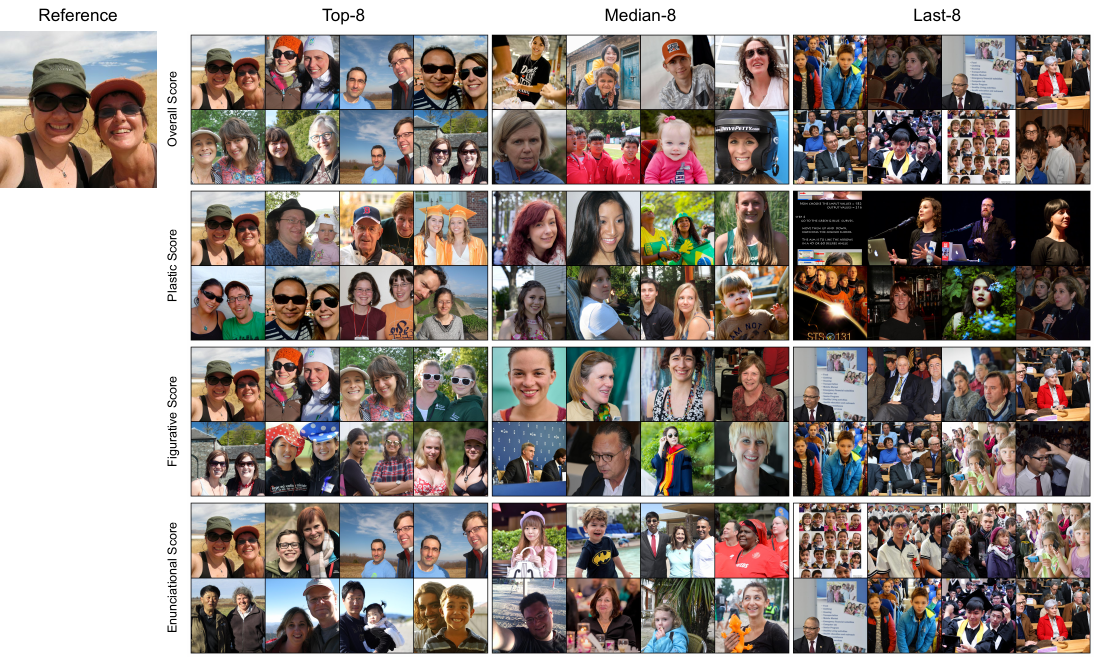}
    \caption{Ranked images using the three scores derived from the highest levels of analysis and the overall score on a reference image including multiple people. Each group of images (Top-8) shows different common aspects such as colors and spatial dispositions of forms (plastic score), person characteristics such as age, gender, emotions, accessories (figurative score) and distances from camera, head/gaze directions while not caring about other details such as gender, emotions an so forth (enunciational score).}
    \label{fig:fresco-retrieve-levels}
\end{figure*}

\begin{figure*}[!tbh]
    \centering
    \includegraphics[width=\textwidth]{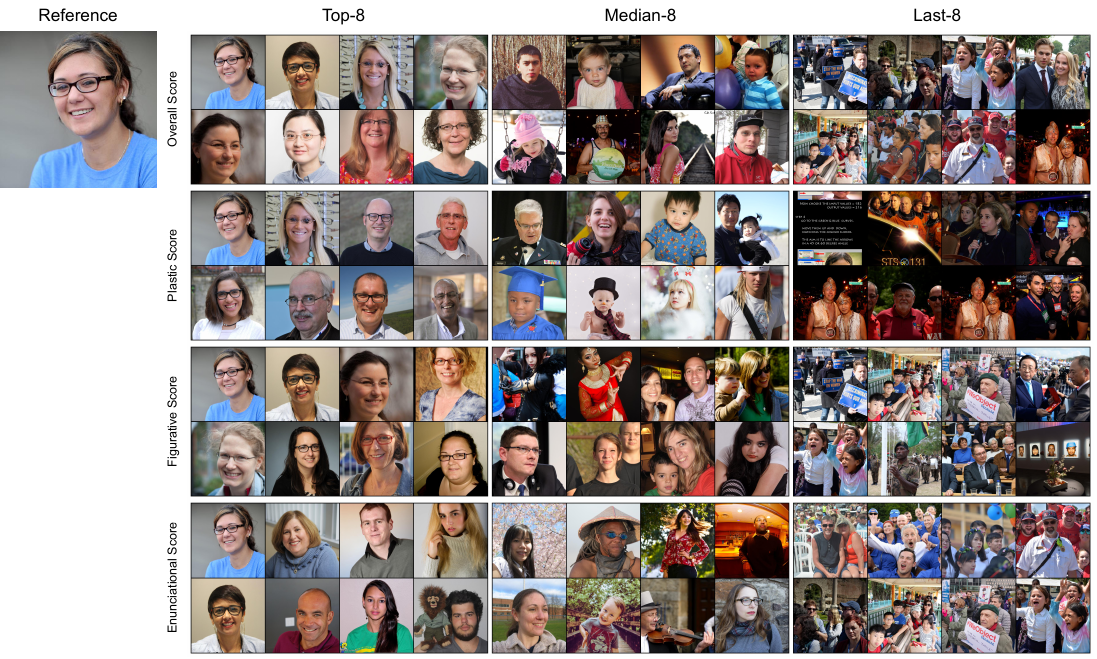}
    \caption{Ranked images using the three scores derived from the highest levels of analysis and the overall score on a reference image including a single person. Each group of images (Top-8) shows different common aspects such as colors and spatial dispositions of forms (plastic score), person characteristics such as age, gender, emotions, accessories (figurative score) and distances from camera, head/gaze directions while not caring about other details such as gender, emotions an so forth (enunciational score).}
    \label{fig:fresco-retrieve-levels2}
\end{figure*}

The FRESCO Score can be employed to rank images based on their similarity. Images can be compared using measures at different levels of aggregation enabling comparisons at varying degrees of detail, from the more general (Overall Score) to specific groups of aspects pertaining to the three levels of analysis (Plastic, Figurative and Enunciational Score) or even down to the more fine-grained characteristics, such as expressions, ethnicity, and head/gaze orientation of each single person depicted in the image, through measures at the lowest level of the FRESCO Score hierarchy. Figure \ref{fig:fresco-retrieve-levels} shows an example of ranking using the Score at the highest level and the three main levels of analysis. In this case, the reference image is compared with the entire FFHQ in-the-wild validation set consisting of 10,000 images. The retrieved groups of images highlight that the Plastic Score, which includes Chromatic (1.2) and Topological categories (1.3), is more susceptible to colors variations (the more distant samples are in general darker) and spatial dispositions of forms (both in terms of covered pixels and distance from camera). The Figurative Score, which covers analysis on characteristics of each person (2.2.2) including among others emotions, gender, and face attributes, allows to retrieve images of people with similar characteristics. In fact, all the images in the Top-8 contain two women of comparable age posing in an outdoor environment, and the presence of accessories such as hats, glasses and necklaces is generally more consistent. The Enunciational Score is instead sensible to the mean distance from camera (3.1.1), the scene (3.1.3) and the direction of head/gaze (3.2.1, 3.2.3) (regardless of the precise spatial position of the subject) and other aspects such as gender, accessories, and so forth. The Overall Score combines the previous scores; when all scores are equally weighted, it can grasp the general content, but it can lose sensitivity to specific aspects. The weight of each level can be properly adjusted to emphasize certain characteristics in the retrieved images, depending on the interests and goals of the scholar using the platform. The number of people (2.2.1.1) is in general well preserved by all scores, due to the effect of the matching strategy that penalizes the presence of unpaired objects as stated in Section \ref{sec:fresco-score}. Indeed, in all cases, the Last-8 images represent in general crowded scenes. The same considerations are also valid for images of a single person as shown in Figure \ref{fig:fresco-retrieve-levels2}.

\begin{figure*}[!tbh]
    \centering
    \includegraphics[width=\textwidth]{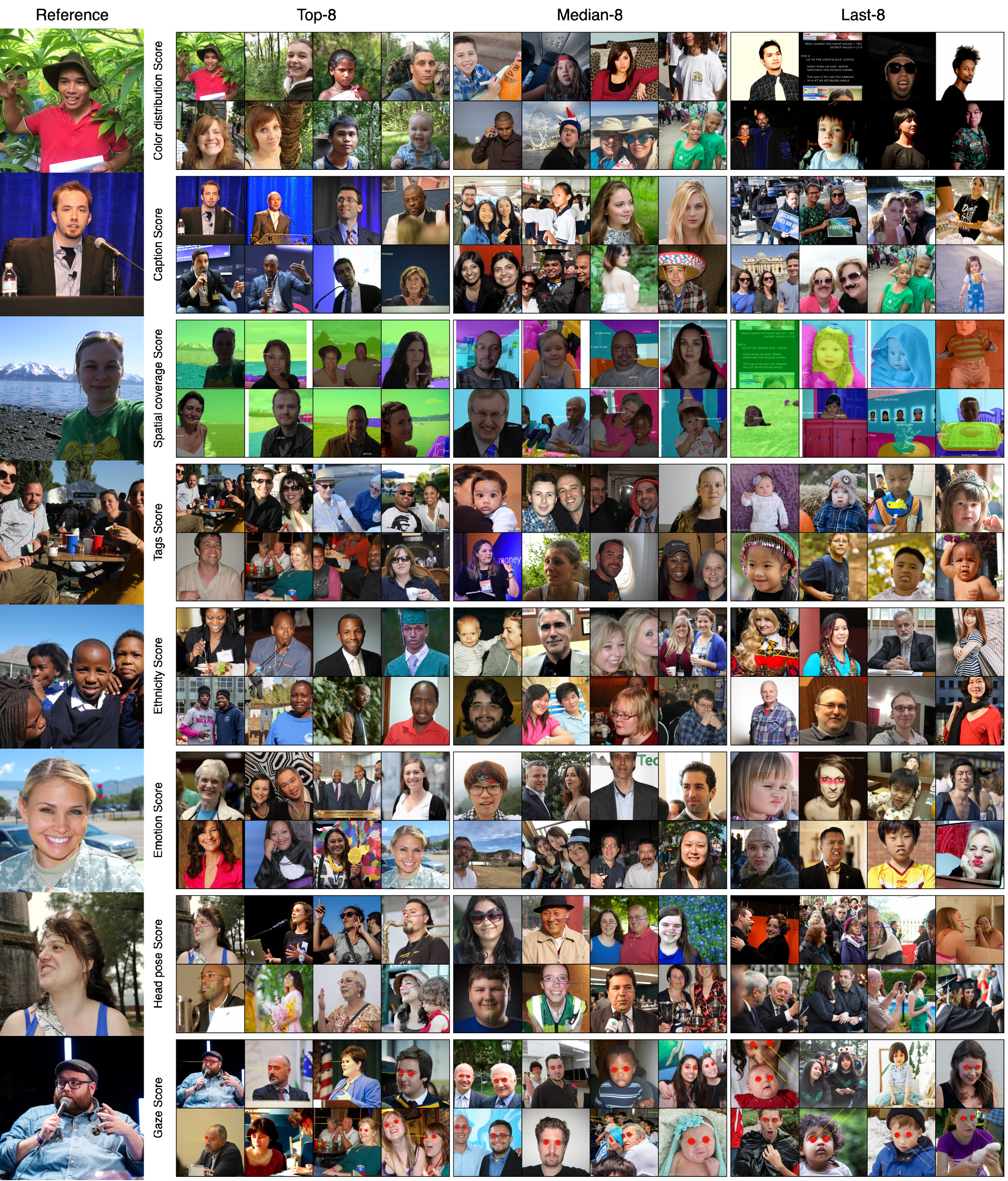}
    \caption{Ranked images using the scores derived from the lowest levels of analysis on a various set of reference images including single and multiple people. Each group of images (Top-8) shows different common aspects strictly related to the considered analysis. For a better interpretability of the results, in this specific test unpaired objects were excluded from the analysis, hence all scores referring to face aspects consider only faces for which a comparable instance can be found among both images, neglecting all the others.}
    \label{fig:fresco-retrieve-single}
\end{figure*}

The ranking based on a subset of single analysis is illustrated in Figure \ref{fig:fresco-retrieve-single}. Each score allows us to find images that are comparable in that specific aspect. Given a reference image, each row shows images that are closest (or farthest) in terms of Color distribution (1.2.1), Textual description (2.4.1), Spatial coverage (1.3.5), and General topics (2.1.1), and in terms of characteristics of single faces including Ethnicity (2.2.2.5), Emotion (2.5.2), Head pose (3.2.1), and Gaze direction (3.2.3). For this specific test, the unpaired objects were excluded from analysis, hence the comparison on faces involves only those who can be directly matched to a comparable one in the second image. Working directly with distances on the models outputs, the Top-8 and the Last-8 images are exactly at the opposite for each specific analysis (within the limit of variability covered by the FFHQ-in-the-wild Validation Set) with the Median-8 in the middle of the distribution, as is definitely evident for head and gaze angles.

\section{Discussion}
\label{sec:discussion}

In this work, we thoroughly explore the application of structural visual semiotics principles to develop a detailed computational framework that facilitates the analysis of large-scale image archives.  In this way, semioticians, as well as scholars in the social sciences and humanities in general, can leverage recent advances in computer vision, and particularly the availability of general purpose models pre-trained on vast amounts of data, also known as foundation models \citep{zhou2023comprehensive}. In the context of social media, for instance, FRESCO can be used to answer questions such as: when self-representing themselves, do people want to show themselves happy or do they prefer to show other moods? Is their face the focus of their images? Do people usually show themselves alone or in company? and many other questions as discussed in \citep{santangelo2023}. 

The outcome of our research is a computational platform that can serve several purposes. First, it converts an image collection from unstructured image data to structured data in tabular format to support the application of data analytics tools \citep{santangelo2023}. This tabular format summarizes to what extent the different traits commonly employed by semioticians to characterize images are expressed by each image, thus configuring a sort of\textit{digital identikit} of an image. Second, it supports a content-based image retrieval system that is based on the plastic, figurative, and/or enunciational content of images, using a configurable similarity score (FRESCO-Score). That said, it is also possible to search for similar images (content-based image retrieval) or to group images (clustering) based on individual characteristics such as caption, human pose, spatial composition, or color distribution. This enables the discovery of similarities and groupings in the data that might otherwise remain unnoticed by researchers \citep{mannisto2022automatic}. \textcolor{black}{Lastly, FRESCO could be used to analyze the quality and content of synthetic images produced by generative models, e.g., to measure alignment with specific instructions, or to quantify systematic differences introduced by generative models with respect to natural images \citep{piano2024latent,piano2024harnessing,barattin2023attribute,otani2023toward}. }

The experimental validation in Section \ref{sec:results}, along with the standalone performance in Table \ref{tab:implementation}, highlights how the extracted features are, in general, of high quality. The agreement between models is generally good, and the distributions extracted from the two datasets analyzed are consistent with how they were sourced and collected. However, there are still challenges and limitations associated with the current implementation of the FRESCO pipeline. Computer vision techniques, although increasingly accurate, may inject various types of errors. Algorithms included in the FRESCO pipeline were selected based on their performance, but noise in the form of errors or uncertainties can arise due to factors such as variations in lighting conditions, image quality, or the complexity of the subject matter. Agreement between different models can be used to filter out uncertain and possible erroneous output. Care must be taken when comparing findings across datasets, but as long as errors can be expected to occur at approximately the same rate on each dataset, a relative comparison can be more reliably estimated than an absolute value \citep{mannisto2022automatic}.

Another challenge that arises when using deep neural networks, especially those relying on a closed set of labels, are out-of-distribution samples. Deep learning models targeting the human face and body, such as for keypoint detection or classification of facial expression, are less prone to this drawback; on the other hand, networks that perform complex and potentially ambiguous multi-label classification, such as scene classification, should be interpreted with greater caution.  In the future, we plan to expand the FRESCO implementation with out-of-distribution detection \citep{recalcati2023toward,yang2024generalized,fort2021exploring} or one-shot domain adaption \citep{yang2024generalized,d2020one}. 

Besides such practical issues, there are also a few limitations that derive from the current architecture of FRESCO. In the field of structural visual semiotics, on which FRESCO foundations were established, the meaning of an image depends not only on what is present and can be seen but also on what it is omitted. Just as we are able to understand the meaning of a sign such as “l” (the letter “L” of the alphabet) as different from “t”, because essentially in the long vertical bar we see there is a short horizontal one at the top missing \citep{de1989cours}, in the same way, we understand that the meaning of a censored image is that subjects seen in other images are not represented. If we look at the famous painting by Manet's \quoten{Olympia}, we see that in her nudity she looks at us proudly, directly and from above, instead of from below, with a more demure and indirect gaze, surrounded by a maid and a cat, instead of a governess and a dog, as in Titian's painting titled \quoten{Venus of Urbino}. Manet's work evidently wants to differentiate itself from Titian's one and from a certain tradition in the representation of naked women who are aware to be observed from men \citep{berger2008ways}, hence we realize that its significance depends, precisely, on the absence of some very significant elements of Titian's own painting and the presence of other deliberately different ones. To overcome this critical gap, FRESCO should be extended with the ability to select, attend, and reason about external and commonsense knowledge \citep{joshi2024contextualizing,ye2018advise}.

\section{Conclusions}
\label{sec:conclusions}
In this study, we extensively investigate the use of structural visual semiotics principles to create FRESCO, a comprehensive computational framework that supports scholars in the analysis of large-scale image archives by leveraging the power of foundational models. In constructing FRESCO, instead of deploying a makeshift collection of deep neural networks, we aimed to represent each category of semiotics through numerical values that can be derived using cutting-edge computer vision models.

We expect FRESCO to further promote the adoption of quantitative methods in visual semiotics \citep{manovich2020cultural}, closing the gap with respect to the analysis of text corpora. At the same time, we hope that FRESCO can foster the interdisciplinary collaboration between computer vision scientists and humanities scholars \citep{ai4mediassh2022}. 

The present study has focused on the technical characteristics of the FRESCO pipeline, outlining its design principle compared to previous studies \citep{mannisto2022automatic}. We also performed internal validation with the aim of investigating the consistency and usability of different extracted characteristics. Currently, we are planning to apply FRESCO to selected case studies involving real-life image collections. In future studies, our aim is to further expand FRESCO by expanding the set of characteristics computed. FRESCO should also be extended to identify and connect elements that are found in an image with those that are absent, but are nonetheless connected to it. To this aim, FRESCO should be integrated with the ability to integrate external and commonsense knowledge, either in the form of structured Knowledge Graphs and/or embedded in Multimodal Large Language Models. 
Finally, further directions include making the computational pipeline more robust, including out-of-distribution detection, as well as adapting and validating the pipeline in other types of image archives, such as historical photography, advertisements, and paintings.

\section*{Acknowledgments}
The present research is funded by the European Research Council (ERC) under the European Union’s Horizon 2020 research and innovation programme (Grant Agreement No 819649-FACETS; PI: Massimo LEONE). The authors wish to thank Pietro Recalcati, Marco Porro, and Enrico Clemente for contributing to the system development. The authors also thank the entire FACETS research team for the stimulating discussions.

\bibliographystyle{model2-names}
\bibliography{main}

\end{document}